\documentclass[9.5pt,journal,final,finalsubmission,twocolumn]{IEEEtran}

\usepackage{graphicx,amsmath,amsfonts}
\usepackage{cite,array,mdwtab,mdwmath}
\usepackage{color}

\def\x{{\vec{x}}}
\def\y{{\vec{y}}}

\def\calW{{\cal W}}

\newcommand{\Real}{\mathbb{R}}

\def\decide#1#2{
  \mathrel{
    \mathop{
      \begin{array}{c}
        >\vspace{-1.4ex}\\<
      \end{array}
      }\limits_{#2}\limits^{#1}
    }
  }

\begin{document}
\title{Behavior Subtraction}
%

\author{Pierre-Marc~Jodoin,~\IEEEmembership{Member,~IEEE,}
        Venkatesh~Saligrama,~\IEEEmembership{Senior~Member,~IEEE,}
        and~Janusz~Konrad,~\IEEEmembership{Fellow,~IEEE}%
\thanks{Manuscript submitted in August 2009}
\thanks{P-M. Jodoin is with the Universit\'e de Sherbrooke,
  D\'epartement d'informatique, Facult\'e des Sciences, 2500 Boul. de
  l'Universit\'e,Sherbrooke, QC, Canada, J1K-2R1,
  pierre-marc.jodoin@usherbrooke.ca}
\thanks{V. Saligrama and J. Konrad are with Boston University,
  Department of Electrical and Computer Engineering, 8 Saint Mary's
  St., Boston, MA 02215 ([srv,jkonrad]@bu.edu)}}


\maketitle

\begin{abstract}

  Background subtraction has been a driving engine for many computer vision and
  video analytics tasks. Although its many variants exist, they all share the
  underlying assumption that photometric scene properties are either static or
  exhibit temporal stationarity. While this works in some applications, the
  model fails when one is interested in discovering {\it changes in scene
    dynamics} rather than those in a static background; detection of unusual
  pedestrian and motor traffic patterns is but one example. We propose a new
  model and computational framework that address this failure by considering
  stationary scene dynamics as a ``background'' with which observed scene
  dynamics are compared. Central to our approach is the concept of an {\it
    event}, that we define as short-term scene dynamics captured over a time
  window at a specific spatial location in the camera field of view. We compute
  events by time-aggregating motion labels, obtained by background subtraction,
  as well as object descriptors (e.g., object size).  Subsequently, we
  characterize events probabilistically, but use a low-memory, low-complexity
  surrogates in practical implementation. Using these surrogates amounts to
  {\it behavior subtraction}, a new algorithm with some surprising properties.
  As demonstrated here, behavior subtraction is an effective tool in anomaly
  detection and localization.  It is resilient to spurious background motion,
  such as one due to camera jitter, and is content-blind, i.e., it works
  equally well on humans, cars, animals, and other objects in both uncluttered
  and highly-cluttered scenes. Clearly, treating video as a collection of
  events rather than colored pixels opens new possibilities for video
  analytics.


\end{abstract}


\section{Introduction}

\PARstart{M}{any} computer vision and video analytics algorithms rely on
background subtraction as the engine of choice for detecting areas of interest
(change). Although a number of models have been developed for background
subtraction, from single Gaussian \cite{Wren97} and mixture of Gaussians
\cite{Stau00} to non-parametric kernel methods \cite{Elga02}\footnote{Although
  models that account for spatial relationships in background subtraction are
  known, their discussion is beyond the scope of this paper.}, they all share
the underlying assumption that photometric scene properties (e.g., luminance,
color) are either static or exhibit temporal stationarity. The static
background assumption works quite well for some applications, e.g., indoor
scenes under constant illumination, while the temporally-stationary background
assumption is needed in other cases, such as outdoor scenes with natural
phenomena (e.g., fluttering leaves).  However, both models fail if one is
interested in discovering {\it changes in scene dynamics} rather than those
taking place in a static background.  Examples of such scenario are: detection
of unusual motor traffic patterns (e.g., too fast or too slow), detection of a
moving group of individuals where a single walking person is expected,
detection of a moving object against shimmering or turbulent water surface
(background motion).  Although each of these challenges can be addressed by a
custom-built method, e.g., explicitly estimating object trajectories or
discovering the number of moving objects, there is no approach to-date that can
address {\it all} such scenarios in a single framework.

In order to address this challenge, instead of searching for photometric
deviations in time, one should look for dynamic deviations in time. To date,
the problem has been attacked primarily by analyzing two-dimensional motion
paths resulting from tracking objects or people
\cite{Hu04,Johnson96,saleemi09,Sumpter00,Stau00}.  Usually, reference motion
paths are computed from a training video sequence first. Then, the same
tracking algorithm is applied to an observed video sequence, and the resulting
paths are compared with the reference motion paths.  Unfortunately, such
methods require many computing stages, from low-level detection to high-level
inferencing \cite{Hu04}, and often result in failure due to multiple,
sequential steps.

In this paper, we propose a new model and computational framework that extend
background subtraction to, what we call, {\it behavior subtraction}
\cite{Jodo08vcip}, while at the same time addressing deficiencies of
motion-path-based algorithms. Whereas in background subtraction static or
stationary photometric properties (e.g., luminance or color) are assumed as the
background image, we propose to use stationary scene dynamics as a
``background'' activity with which observed scene dynamics are compared. The
approach we propose requires neither computation of motion nor object tracking,
and, as such, is less prone to failure. Central to our approach is the concept
of an {\it event}, that we define as short-term scene dynamics captured over a
time window at a specific spatial location in the camera field of view. We
compute events by time-aggregating motion labels and/or suitable object
descriptors (e.g., size).  Subsequently, we characterize events
probabilistically as random variables that are independent and identically
distributed ({\it iid}) in time.
%
%
Since the estimation of a probability density function (PDF) at each location
is both memory- and CPU-intensive, in practical implementation we resort to a
low-memory, low-complexity surrogate. Using such a surrogate amounts to
behavior subtraction, a new algorithm with some surprising properties. As we
demonstrate experimentally, behavior subtraction is an effective tool in
anomaly detection, including localization, but can also serve as motion
detector very resilient to spurious background motion, e.g., resulting from
camera jitter.  Furthermore, it is content-blind, i.e., applicable to humans,
cars, animals, and other objects in both uncluttered and highly-cluttered
scenes.

%

This paper is organized as follows. In Section~\ref{sec:previous}, we review
previous work. In Section~\ref{sec:backsub}, we recall background subtraction
and introduce notation. In Section~\ref{sec:behspace}, we introduce behavior
space and the notion of an event, while in Section~\ref{sec:behsubframework} we
describe the behavior subtraction framework. In Section~\ref{sec:expres}, we
discuss our experimental results and in Section~\ref{sec:concl} we draw
conclusions.

\section{Previous work}
\label{sec:previous}

There are two fundamental approaches to anomaly detection. One approach is to
explicitly model all anomalies of interest, thus constructing a dictionary of
anomalies, and for each observed video to check if a match in the dictionary
can be found. This is a typical case of classification, and requires that {\it
  all} anomaly types be known {\em a priori}. Although feasible in very
constrained scenarios, such as detecting people carrying
boxes/suitacases/handbags \cite{Chuang09}, detecting abandoned objects
\cite{Smith06} or identifying specific crowd behavior anomalies
\cite{Mehran09}, in general this approach is not practical for its inability to
deal with unknown anomalies.

An alternative approach is to model normality and then detect deviations from
it.  In this case, no dictionary of anomalies is needed but defining and
modeling what constitutes normality is a very difficult task.  One way of
dealing with this difficulty is by applying machine learning that automatically
models normal activity based on some training video. Then, any monitored
activity different from the normal pattern is labeled as anomaly.  A number of
methods have been developed that apply learning to two-dimensional motion paths
resulting from tracking of objects or people \cite{Hu04}. Typically, the
approach is implemented in two steps.  In the first step, a large number of
``normal'' individuals or objects are tracked over time.  The resulting paths
are then summarized by a set of motion trajectories, often translated into a
symbolic representation of the background activity. In the second step, new
paths are extracted from the monitored video and compared to those computed in
the training phase.

Whether one models anomaly or normality, the background activity must be
somehow captured. One common approach is through graphical state-based
representations, such as hidden Markov models or Bayesian networks
\cite{Hu04,Kumar05,Bennewitz03,Oliver00,Vaswani05}. To the best of our
knowledge Johnson and Hogg \cite{Johnson96} were the first to consider human
trajectories in this context. The method begins by vector-quantizing tracks and
clustering the result into a predetermined number of PDFs using a neural
network. Based on the training data, the method predicts trajectory of a
pedestrian and decides if it is anomalous or not. This approach was
subsequently improved by simplifying the training step \cite{Sumpter00} and
embedding it into a hierarchical structure based on co-occurrence statistics
\cite{Stau00}.  More recently, Saleemi {\em et al.} \cite{saleemi09} proposed a
stochastic, non-parametric method for modeling scene tracks.  The authors claim
that the use of predicted trajectories and tracking method robust to occlusions
jointly permit the analysis of more general scenes, unlike other methods that
are limited to roads and walkways.

Although there are advantages to using paths as motion features, there are
clear disadvantages as well. First, tracking is a difficult task, especially in
real time.  Since the anomaly detection is directly related to the quality of
tracking, a tracking error will inevitably bias the detection step.  Secondly,
since each individual or object monitored is related to a single path, it is
hard to deal with people occluding each other. For this reason, path-based
methods aren't well suited to highly-cluttered environments.

Recently, a number of anomaly detection methods have been proposed that do not
use tracking. These methods work at pixel level and use either motion vectors
\cite{Kim09,Dong09,Adam08} or motion labels \cite{Cui07,Xiang06,Oh03} to
describe activity in the scene.  They all store motion features in an
image-like 2D structure (be it probabilistic or not) thus easing memory and CPU
requirements. For example, Xiang {\em et al.} \cite{Xiang06} represent moving
objects by their position, size, temporal gradient and the so-called ``pixel
history change'' (PHC) image that accumulates temporal intensity differences.
During the training phase, an EM-based algorithm is used to cluster the moving
blobs, while at run-time each moving object is compared to the pre-calculated
clusters.  The outlying objects are labeled as anomalous. Although the concept
of PHC image is somewhat similar to the behavior image proposed here, Xiang
{\it et al.} do not use it for anomaly detection but for identification of
regions of interest to be further processed.


%

A somewhat different approach using spatio-temporal intensity correlation has
been proposed by Shechtman and Irani \cite{Boiman07}. Here, an observed
sequence is built from spatio-temporal segments extracted from a training
sequence. In this analysis-by-synthesis method, only regions that can be built
from large contiguous chunks of the training data are considered normal.


Our approach falls into the category of methods that model normality and look
for outliers, however it is not based on motion paths but on simple pixel
attributes instead. Thus, it avoids the pitfalls of tracking while affording
explicit modeling of normality at low memory and CPU requirements. Our
contributions are as follows. We introduce the concept of an event, or
short-term scene dynamics captured over a time window at a specific spatial
location in the camera field of view. With each event we associate features,
such as size, direction, speed, busy time, color, etc., and propose a
probabilistic model based on time-stationary random process.  Finally, we
develop a simple implementation of this model by using surrogate quantities
that allow low-memory and low-CPU implementation.

\section{Background Subtraction: Anomaly Detection in Photometric Space}
\label{sec:backsub}

We assume in this paper that the monitored video is captured by a fixed camera
(no PTZ functionality) that at most undergoes jitter, e.g., due to wind load or
other external factors.

Let $\vec{I}$ denote a color video sequence with $\vec{I}_t(\x)$ denoting color
attributes (e.g., $R,G,B$) at specific spatial location $\x$ and time $t$. We
assume that $\vec{I}_t(\x)$ is spatially sampled on 2-D lattice $\Lambda$,
i.e., $\x\in\Lambda\subset R^2$ is a pixel location.  We also assume that it is
sampled temporally, i.e., $t=k\Delta t$, $k\in Z$, where $\Delta t$ is the
temporal sampling period dependent on the frame rate at which the camera
operates. For simplicity, we assume $\Delta=1$ in this paper, i.e., normalized
time. We denote by $\vec{I}_t$ a frame, i.e., a restriction of video $\vec{I}$
to specific time $t$.

In traditional video analysis, color and luminance are pivotal quantities in
the processing chain. For example, in background subtraction, the driving
engine of many video analysis tasks, the color of the background is assumed
either static or stationary. Although simple frame subtraction followed by
thresholding may sometimes suffice in the static case, unfortunately it often
fails due to acquisition noise or illumination changes. If the background
includes spurious motion, such as environmental effects (e.g., rain, snow),
fluttering tree leaves, or shimmering water, then determining outliers based on
frame differences is insufficient. A significant improvement is obtained by
determining outliers based on PDF estimates of features such as color.  Assume
that $P_{RGB}$ is a joint PDF of the three color components estimated using a
3-D variant of the mixture-of-Gaussians model \cite{Stau00} or the
non-parametric model \cite{Elga02} applied to a training video sequence.
$P_{RGB}$ can be used to test if a color at specific pixel and time in the
monitored video is sufficiently probable, i.e., if $P_{RGB}(\vec{I}_t(\x)) >
\tau$, where $\tau$ is a scalar threshold, then $\vec{I}_t(\x)$ is likely to be
part of the modeled background, otherwise it is deemed moving.

Although the thresholding of a PDF is more effective than the thresholding of
frame differences, it is still executed in the space of photometric quantities
(color, luminance, etc.), and thus unable to directly account for scene
dynamics.  However, modeling of background dynamics (activities) in the
photometric space is very challenging. We propose an alternative that is both
conceptually simple and computationally efficient. First, we remove the
photometric component by applying background subtraction and learn the
underlying stationary statistical characterization of scene dynamics based on a
two-state (moving/static) renewal model. Then, we reliably infer novelty as a
departure from the normality.

%

%

\section{Behavior Space: From Frames to Events}
\label{sec:behspace}

As color and luminance contain little direct information on scene dynamics, we
depart from this common representation and adopt motion label as our atomic
unit. Let $L_t(\x)$ be a binary random variable embodying the presence of
motion ($L=1$) or its absence ($L=0$) at position $\x$ and time $t$. Let
$l_t(\x)$ be a specific realization of $L_t(\x)$ that can be computed by any of
the methods discussed in Section~\ref{sec:backsub}, or by more advanced methods
accounting for spatial label correlation \cite{Migd05,Shei05,McHu09spl}.

While some of these methods are robust to noise and background activity, such as
rain/snow or fluttering leaves, they often require a large amount of memory and
are computationally intensive. Since simplicity and computational efficiency
are key concerns in our approach, we detect motion by means of a very simple
background subtraction method instead, namely
\begin{eqnarray}
	l_t(\x) = |I_t(\x)-b_t(\x)|>\tau,
	\label{eq:md}
\end{eqnarray}
where $\tau$ is a fixed threshold and $b_t$ is the background image computed as follows
\begin{eqnarray}
	b_{t+1}(\x) = (1-\rho)b_t(\x) + \rho I_t(\x)
\end{eqnarray}
with $\rho$ in the range 0.001-0.01. This linear background update allows to
account for long-term changes. Although this method is sensitive to noise and
background activity, it is trivial to implement, requires very little memory
and processing power, and depends on one parameter only. Clearly, replacing
this method with any of the advanced techniques will only improve the
performance of our approach.

\begin{figure}[tb]

  \centering
  \begin{tabular}{cc}
    \footnotesize Video frame $\vec{I}_{t=t_0}$ &
    \footnotesize Motion label field $l_{t=t_0}$\\
    \includegraphics [width=4cm]{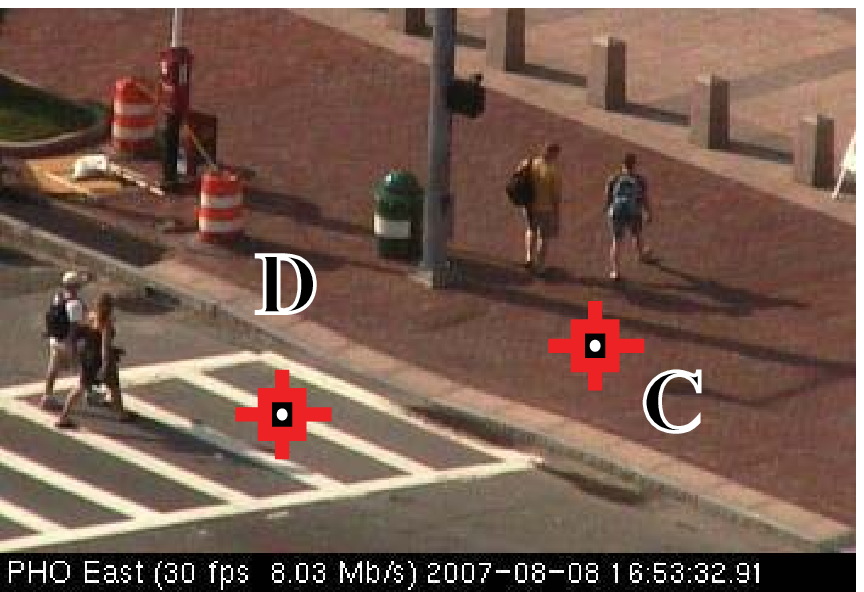} &
    \includegraphics [width=4cm]{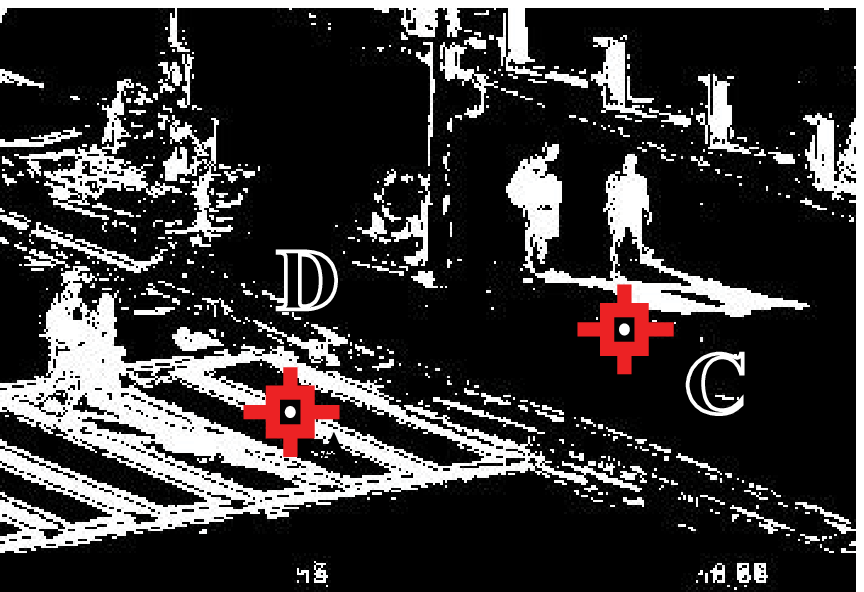}
  \end{tabular}
  \centerline{\small (a)}

  \footnotesize Motion label at pixel ``C'': $l(\x=\vec{x}_C)$\\
  \centerline{\includegraphics [width=8.5cm]{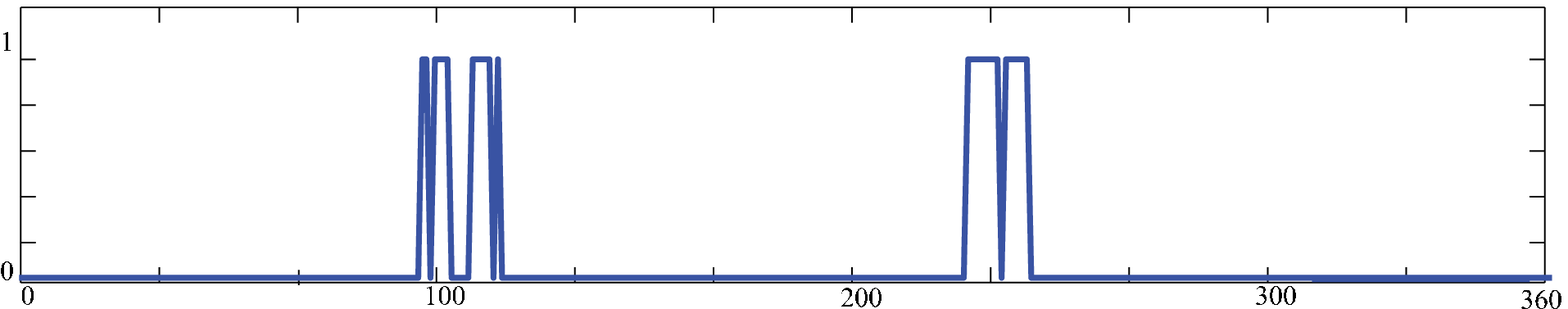}}
  \footnotesize Motion label at pixel ``D'': $l(\x=\vec{x}_D)$\\
  \centerline{\includegraphics [width=8.5cm]{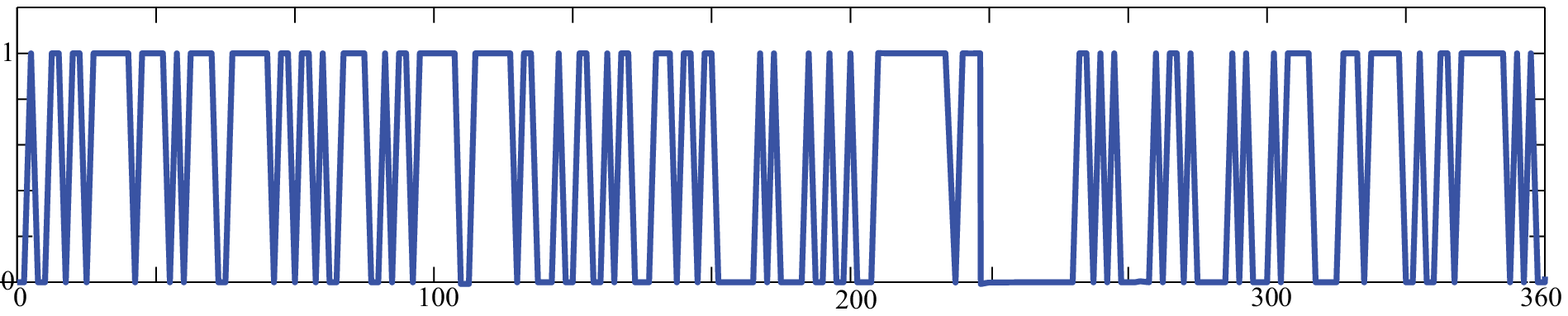}}
  \centerline{\small (b)}
  \medskip

  \footnotesize Behavior signature at pixel ``C'': $f(\x=\vec{x}_C)$\\
  \centerline{\includegraphics [width=8.5cm]{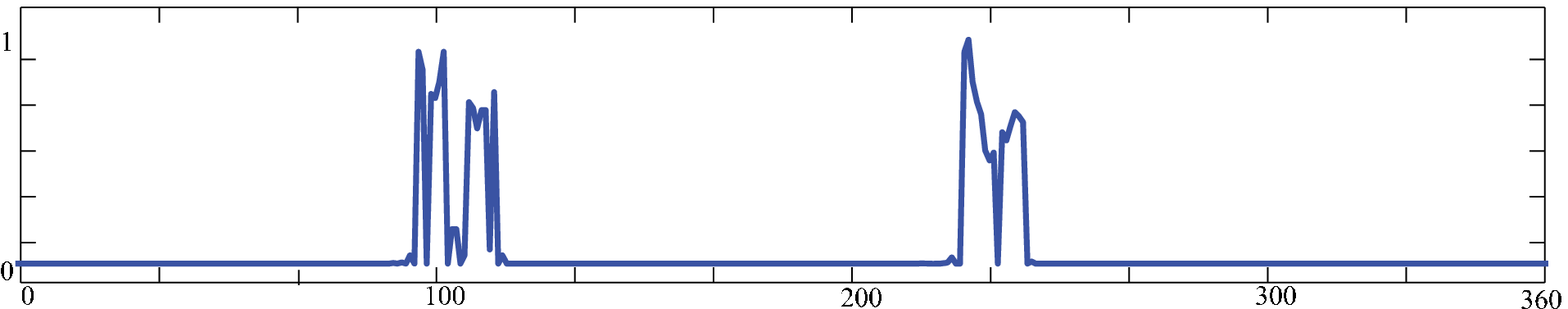}}
  \footnotesize Behavior signature at pixel ``D'': $f(\x=\vec{x}_D)$\\
  \centerline{\includegraphics [width=8.5cm]{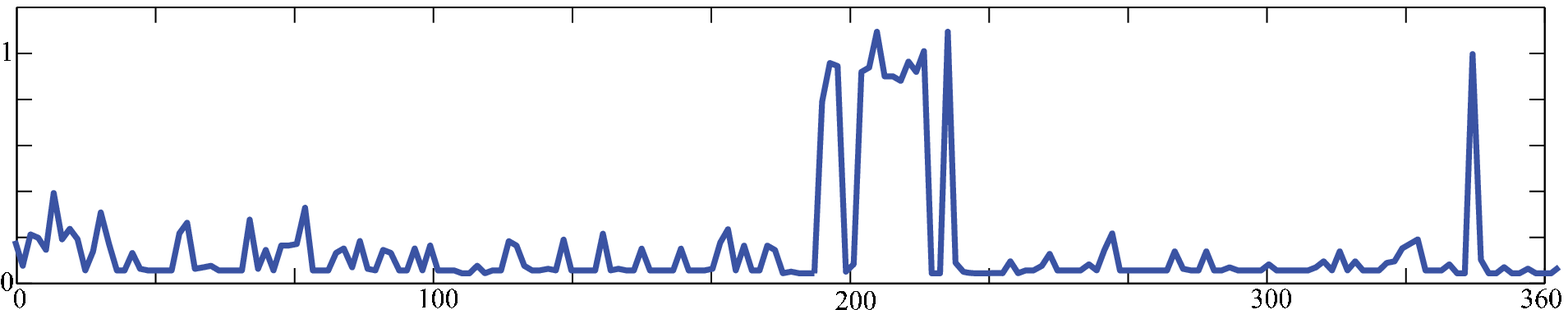}}
  \centerline{\small (c)}


  \caption{\small (a) Video frame $\vec{I}_{t=t_0}$ captured by a vibrating camera
    and the corresponding motion label field $l_{t=t_0}$. (b) Binary waveforms
    showing the time evolution of motion labels $l$ at two locations (marked
    $C$ and $D$ in (a)).  (c) Behavior signatures at the same locations
    computed using the object-size descriptor (\ref{eqn:motobjdes}). The pixel
    located near intensity edge ($D$) is ``busy'', due to camera vibrations,
    compared to the pixel located in a uniform-intensity area ($C$).  The large
    bursts of activity in behavior signatures correspond to pedestrians.}
\label{fig:MDTraining}
\vglue -0.4cm
\end{figure}

Fig.~\ref{fig:MDTraining} shows an example realization of motion label field
$L_t$ computed by the above method as well as a binary waveform showing
temporal evolution of motion label at specific location $\x$
(Fig.~\ref{fig:MDTraining}.b). Each such waveform captures the amount of
activity occurring at a given spatial location during a certain period of time
and thus can be considered as a simple {\it behavior signature}.  For instance,
patterns associated with random activity (fluttering leaves), periodic activity
(highway traffic), bursty activity (sudden vehicle movement after onset of
green light), or no activity, all have a specific behavior signature. Other
behavior signatures than a simple on/off motion label are possible.

\paragraph{Object descriptor}

A moving object leaves a behavior signature that depends on its features such
as size, shape, speed, direction of movement, etc. For example, a large moving
object will leave a wider impulse than a small object
(Fig.~\ref{fig:MDTraining}.b), but this impulse will get narrower as the object
accelerates. One can combine several features in a descriptor in order to make
the behavior signature more unique. In fact, one can even add color/luminance
to this descriptor in order to account for photometric properties as well.
Thus, one can think of events as spatio-temporal units that describe what type
of activity occurs and also what the moving object looks like.

Let a random variable $F$ embody object description\footnote{$F$ is a random
  vector if the descriptor includes multiple features.}, with $f$ being its
realization.  In this paper, we concentrate on object descriptor based on
moving object's size for two reasons. First, we found that despite its
simplicity it performs well on a wide range of video material (motor traffic,
pedestrians, objects on water, etc.); it seems the moving object size is a
sufficiently discriminative characteristic. Secondly, the size descriptor can
be efficiently approximated as follows:
\begin{eqnarray}\label{eqn:motobjdes}
  f_t(\x) = \frac{1}{N\times N}
            \sum_{\y\in {\cal N}(\x); \y\Join\x}
            \delta(l_t(\x),l_t(\y)),
\label{eq:motionFeature}
\end{eqnarray}
where ${\cal N}(\x)$ is an $N\times N$ window centered at $\x$ and $\y\Join\x$
means that $\y$ and $\x$ are connected (are within the same connected
component).  $\delta(\cdot) = 1$ if and only if $l_t(\x)=l_t(\y)=1$, i.e., if
both $\x$ and $\y$ are deemed moving, otherwise $\delta(\cdot)=0$. Note that
$f_t(\x)=0$ whenever $l_t(\x)=0$. This descriptor is zero for a pixel away from
the object, increases non-linearly as the pixel moves closer to the object and
saturates at 1.0 for pixels inside a large object fully covering the window
$\cal N$.

Fig.~\ref{fig:MDTraining}.c shows an example of behavior signature based on the
size descriptor. Clearly, $f_t(\x)=0$ means inactivity while $f_t(\x)>0$ means
activity caused by a moving object; the larger the object, the larger the
$f_t(\x)$ until it saturates at 1. The video frame shown has been captured by a
vibrating camera and thus a noisy behavior signature for pixel ``D'' that is
close to an intensity edge.

\paragraph{Event model}

\begin{figure}[tb]
  \centering
  \includegraphics[width=6.5cm]{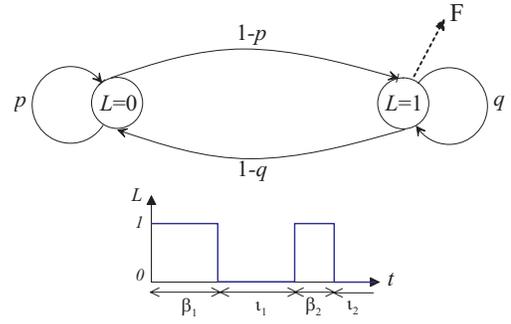}
\vglue -0.2cm
  \caption{Markov chain model for dynamic event $E$: $p,q$ are state
    probabilities (static and moving, respectively), and $1-p,1-q$ are
    transition probabilities. $\beta_1,\iota_1,\beta_2,\iota_2$ denote
    consecutive busy and idle intervals. With each busy interval is associated
    an object descriptor $F$, such as its size, speed/direction of motion,
    color, luminance, etc.}
\label{fig:event_model}
\vglue -0.4cm
\end{figure}

An event needs to be associated with a time scale. For example, a short time
scale is required to capture an illegal U-turn of a car, whereas a long time
scale is required to capture a traffic jam. We define an event $E_t(\x)$ for
pixel at $\x$ as the behavior signature (object size, speed, direction as the
function of time $t$) left by moving objects over a $w$-frame time window, and
model it by a Markov model shown in Fig.~\ref{fig:event_model}.

For now, consider only the presence/absence of activity ($L$) as the object
descriptor. Assuming $\pi$ to be the initial busy-state probability ($L=1$),
the probability of sequence $\{L_i=l_i\}_\calW =
(l_{t-w+1}(\x),\,l_{t-w+2}(\x),\,\ldots,\,l_t(\x))$, at location $\x$ and
within the time window $\calW=[t-w+1,t]$, can be written as follows:
\begin{eqnarray}\label{eqn:event_prob}
  P_{\x}(\{L_i=l_i\}_\calW)
    &=& \pi q^{\beta_1} (1-q) p^{\iota_1} (1-p) q^{\beta_2} (1-q)
        p^{\iota_2} ...\nonumber\\
    &=& \pi q^{\sum_k \beta_k} p^{\sum_k \iota_k} (1-q)^m (1-p)^n\\
    &=& \pi (q/p)^{\sum_k \beta_k} p^w (1-q)^m (1-p)^n,\nonumber
\end{eqnarray}
%
where the binary sequence of 0's and 1's is implicitly expressed through the
busy intervals $\beta_k$ (Fig.~\ref{fig:event_model}).  Note that $m,n$ are the
numbers of transitions ``moving $\rightarrow$ static" and "static $\rightarrow$
moving", respectively. The last line in (\ref{eqn:event_prob}) stems from the
fact that the sum of busy and idle intervals equals the length of time window
$\calW$.  This expression can be simplified by taking negative logarithm:
\begin{eqnarray}\label{eqn:event_prob_log}
  -\log P_\x(\{L_i=l_i\}_\calW) &=& -\log\pi - (\log q/p)\sum_k \beta_k - w \log p -\nonumber\\
                  & & m \log (1-q) - n \log (1-p),\\
                  &=& A_0 + A_1 \sum_{k=t-w+1}^t l_k(\x) + A_2 \kappa_t(\x),\nonumber
\end{eqnarray}
where $A_0,A_1,A_2$ are constants, the second term measures the total busy time
using motion labels and $\kappa_t(\x)$ is proportional to the total number of
transitions in time window $\calW$ at $\x$.

Thus far we have assumed that the moving object was described only by motion
labels $L_t(\x)$. Suppose now that also a descriptor $F_t(\x)$, such as the
size, is associated with the moving object at location $\x$ and
time $t$ within a busy period in time window $\calW$, i.e.,
$t\in\beta_k\subset\calW$.  The random variable (vector) $F_t(\x)$ is described
by a conditional distribution dependent on the state of the Markov process, as
illustrated in Fig.~\ref{fig:event_model}. We assume that $F_t(\x)$ is
conditionally independent of other random variables $F_{t_0}(\x), t_0\neq t$
when conditioned on the underlying state of the Markov process, and that its
distribution has exponential form when busy and point mass when idle:
\begin{eqnarray}\label{eqn:Gibbs_1}
  P_\x(F_t=f_t \mid L_t = k) = \left\{
      \begin{array}{ll}
         \frac{1}{Z_1} e^{-A_3 f_t(\x)}, & k=1,\\
         \delta(0),        & k=0.
      \end{array}\right.
\end{eqnarray}
%
%
where $Z_1$ is a partition function and $\delta$ is the Kronecker delta.  If the
descriptor $F$ includes object size, the above distribution suggests that the
larger the object passing through $\x$ the less likely it is, and also that
with probability 1 it has size zero in idle intervals (consistent with
Fig.~\ref{fig:event_model}). This is motivated by the observation that
small-size detections are usually associated with false positives when
computing $L_t$.  Should $F$ include speed, faster objects would be less
likely, a realistic assumption in urban setting.  The model would have to be
modified should the descriptor include direction of motion (e.g., horizontal
motion more likely for highway surveillance with a suitably-oriented camera) or
luminance/color (e.g., all photometric properties equally likely).

Note that more advanced descriptor models can be incorporated as well. For
instance, one can enforce temporal smoothness of the descriptor (e.g., size)
for object passing through location $\x$ {\it via} a (temporal) Gibbs
distribution with 2-element cliques:
\begin{eqnarray*}
%
%
  \lefteqn{P_\x(\{F_i=f_i\}_\calW \mid L=1 ) = } \\
  && \frac{1}{Z_2} e^{-A_4 \sum_{k: \beta_k\subset\calW}
     \sum_{(j,j+1)\in\beta_k} f_j(\x) f_{j+1}(\x)},
\end{eqnarray*}
where $\{F_i=f_i\}_\calW$ denotes a sequence of descriptors appearing in the
temporal window $\calW$, and $A_4$ is a constant. This model controls temporal
smoothness of the descriptor $F$, and can be used to limit, for example, size
variations in time.  Nevertheless, for simplicity we omit this model in our
further developments.

Combining the descriptor model (\ref{eqn:Gibbs_1}) with the $L$-based event
model (\ref{eqn:event_prob}-\ref{eqn:event_prob_log}) leads to a joint
distribution:
\begin{eqnarray}\label{eqn:fl-jointdist}
  \lefteqn{P_\x(\{L_i=l_i\}_\calW,\{F_i=f_i\}_\calW) =} \nonumber\\
    && P_\x(\{F_i=f_i\}_\calW \mid \{L_i=l_i\}_\calW)\cdot P_\x(\{L_i=l_i\}_\calW) =\\
    && \prod_{i\in\calW} P_\x(F_i=f_i \mid L_i=l_i)\cdot P_\x(\{L_i=l_i\}_\calW)\nonumber
\end{eqnarray}
where the last line stems from the conditional independence of $F_i$'s when
conditioned on $L$'s assumed earlier.
Taking the negative logarithm and using equations (\ref{eqn:event_prob_log})
and (\ref{eqn:Gibbs_1}) results in:
\begin{eqnarray}\label{eqn:event_prob_total}
  \lefteqn{-\log P_\x(\{L_i=l_i\}_\calW,\{F_i=f_i\}_\calW) = A_0^\prime\  + }\\
    && A_1 \sum_{k=t-w+1}^t l_k(\x) +
       A_2 \kappa_t(\x) + A_3 \sum_{k=t-w+1}^t f_k(\x)l_k(\x),\nonumber
\end{eqnarray}
where $A_0^\prime$ accounts for $Z_1$ (\ref{eqn:Gibbs_1}) and the last term is
the sum of descriptors in all busy periods in $\calW$. Note that the constant
$A_2$ is positive, thus reducing the probability when frequent ``moving
$\rightarrow$ static" and "static $\rightarrow$ moving" transitions take place.
The constant $A_1$ may be negative or positive depending on the particular
values of $q$ and $p$ in the Markov model; increasing busy periods within
$\calW$ will lead to an increased ($q>p$) or decreased ($q<p$) joint
probability.

Note that at each location $\x$ the above model implicitly assumes independence
among the busy and idle periods as well as conditional independence of $F_t$
when conditioned on $L_t=l_t$. This assumption is reasonable since different
busy periods at a pixel correspond to different objects while different idle
periods correspond to temporal distances between different objects.  Typically,
these are all independent\footnote{We have performed extensive experiments
  ranging from highway traffic to urban scenarios and the results appear to be
  consistent with these assumptions.}.

With each time $t$ and position $\x$ we associate an event $E_t$ that
represents the statistic described in (\ref{eqn:event_prob_total}), namely,
\begin{eqnarray}
  E_t(\x) = \sum_{k=t-w+1}^t (A_1 L_k(\x) + A_3 F_k(\x)L_k(\x))
            + A_2 {\cal K}_t(\x),
  \label{eqn:event_prob_suff}
\end{eqnarray}
where the constant $A_0^\prime$ was omitted as it does not contribute to the
characterization of dynamic behavior (identical value across all $\x$ and $t$)
and ${\cal K}$ is a random variable associated with realization $\kappa$
(number of transitions).  The main implication of the above event description
is that it serves as a sufficient statistic for determining optimal decision
rules \cite{Poor94}.
%

\paragraph{Anomaly Detection Problem}

We first describe anomaly detection abstractly. We are given data, $\omega \in
\Omega \subset \Real^d$. The nominal data are sampled from a multivariate
density $g_0(\cdot)$ supported on the compact set $\Omega$.  Anomaly
detection~\cite{Zhao09} can be formulated as a composite hypothesis
testing problem. Suppose the test data, $\omega$, come from a mixture
distribution, namely, $f(\cdot) = (1-\xi) g_0(\cdot) + \xi g_1(\cdot)$ where
$g_1(\cdot)$ is also supported on $\Omega$. Anomaly detection involves testing
the following nominal hypothesis
\begin{eqnarray*}
  H_0: \xi = 0\,\,\, \mbox{versus the alternative (anomaly)}\,\,\, H_1 : \xi>0.
\end{eqnarray*}
The goal is to maximize the detection power subject to false alarm level
$\alpha$, namely, $\mbox{Prob}(\mbox{declare } H_1 \mid H_0) \leq \alpha$.
Since the mixing density is unknown, it is usually assumed to be uniform. In
this case the optimal uniformly most powerful test (over all values of $\xi$)
amounts to thresholding the nominal density \cite{Poor94}.
%
%
We choose a threshold $\tau(\alpha)$ and declare the observation, $\omega$, as
an outlier according to the following log-likelihood test:
\begin{eqnarray}\label{eqn:gtest}
  -\log(g_0(\omega)) \decide{H_1}{H_0} \tau(\alpha)
\end{eqnarray}
where $\tau(\alpha)$ is chosen to ensure that the false alarm probability is
smaller than $\alpha$. It follows that such a choice is the uniformly most
powerful decision rule. Now the main problem that arises is that $g_0(\cdot)$
is unknown and has to be learned in some way from the data. The issue is that
$\omega$ could be high-dimensional and learning such distributions may not be
feasible. This is further compounded in video processing by the fact that it is
even unclear what $\omega$, i.e., the features, should be.

It is worth reflecting how we have addressed these issues through our specific
setup. We are given $w$ video frames, $I_{t-w+1},\,I_{t-w+2},\,\ldots,\,I_t$
and a specific location $\x$, and our task is to determine whether this
sequence is consistent with nominal activity or, alternatively, it is
anomalous. We also have training data that describes the nominal activity. In
this context, our Markovian model provides a representation for the observed
video frames. This representation admits a natural factorization, wherein
increasingly complex features can be incorporated, for example through
Markov-Gibbs models.
%
%
Furthermore, the log-likelihood is shown to be reduced to a scalar sufficient
statistic, which is parameterized by a finite set of parameters ($A_j$'s in
(\ref{eqn:event_prob_suff})). Consequently, the issue of learning
high-dimensional distribution is circumvented and one is left with estimating
the finite number of parameters, which can be done efficiently using standard
regression techniques.  The problem of anomaly detection now reduces to
thresholding the event $E_t=e_t$ according to (\ref{eqn:gtest}):
\begin{eqnarray*}
  e_t(\x) \decide{H_1}{H_0}\tau(\alpha),
\end{eqnarray*}
or, explicitly,
\begin{eqnarray}\label{eqn:event_hyptest}
  \sum_{k=t-w+1}^t (A_1 l_k(\x) + A_3 f_k(\x)l_k(\x)) +
  A_2 \kappa_t(\x)\decide{H_1}{H_0}\tau(\alpha).
\end{eqnarray}
Our task is to find an appropriate threshold $\tau(\alpha)$ so that the false
alarms are bounded by $\alpha$. Note that our events are now scalar and
learning the density function of a 1-D random variable can be done efficiently.
The main requirement is that $E_t(\x)$ be a stationary ergodic stochastic
process, which will ensure that the CDF can be accurately estimated:
\begin{eqnarray*}
  \frac{1}{w} \sum_{k=t-w+1}^t {1{\hskip -2.5 pt}\hbox{I}}_{\{E_t(\x)\geq\eta\}}(e_t(\x))
    \longrightarrow\mbox{Prob}_{\x} \{E \geq \eta \},
\end{eqnarray*}
where ${1{\hskip -2.5 pt}\hbox{I}}_{\{E_t(\x)\geq\eta\}}(e_t(\x))$ is an
indicator function, equal to $1$ when $e_t(\x)>\eta$ and $0$ otherwise, while
$\mbox{Prob}_\x$ denotes the representative stationary distribution for $E_t$
at any time $t$. For Markovian processes this type of ergodicity is
standard~\cite{Karl75}. One extreme situation is to choose a threshold that
ensures zero false alarms. This corresponds to choosing $\tau(0) = \max_t e_t$,
i.e., the maximum value of the support of all events in the training data.

\begin{figure}[tb]
  \centering
  \includegraphics[width=7cm]{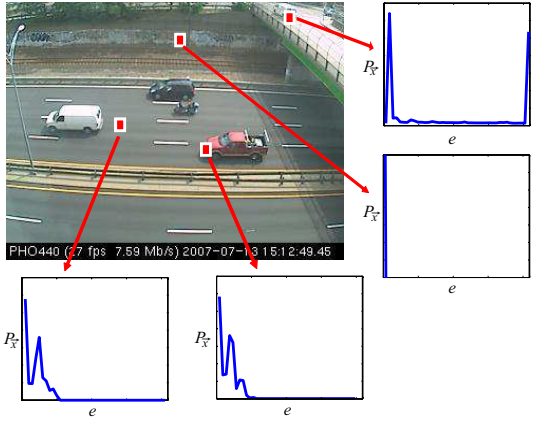}
\vglue -0.2cm
  \caption{Event model PDF estimated for four different
    pixels.  The two pixels in traffic lanes have similar histograms due to the
    fact that their behaviors are very similar (continuous highway traffic).
    The pixel above the traffic is in the idle area of the video, so its
    histogram has a high peak near zero, the pixel on the overpass has a
    bimodal distribution caused by the traffic light.}
\label{fig:histograms}
\vglue -0.5cm
\end{figure}

Although the anomaly detection algorithm we describe in the next section
requires no explicit estimation of the above CDF, it is nevertheless
instructive to understand its properties. Fig.~\ref{fig:histograms} shows
example PDFs for our test statistic $e_t(x)$ estimated from training data using
smoothed histograms. Note different histogram shapes depending on the nature of
local activity.


\section{Behavior Subtraction Framework}
\label{sec:behsubframework}

In the previous section, we presented object and event models, and explained
how they fit into the problem of anomaly detection. In principle, once the
event model is known various statistical techniques can be applied but this
would require significant memory commitment and computational resources.
Below, we propose an alternative that is memory-light and processor-fast and
yet produces very convincing results.



\subsection{Behavior Images}
\label{ssec:behimg}

As mentioned in the previous section, one extreme situation in anomaly
detection is to ensure zero false alarms. This requires a suitable threshold,
namely $\tau(0) = \max_t e_t$, equal to the maximum value of the support of all
events in the training data. This threshold is space-variant and can be
captured by a 2-D array:
\begin{eqnarray}\label{eqn:bbehimg1}
  B(\x) = \max_{t\in[1,M]} e_t(\x),
\end{eqnarray}
where $M$ is the length of the training sequence. We call $B$ the {\em
  background behavior image} \cite{Jodo08vcip} as it captures the background
activity (in the training data) in a low-dimension representation (one scalar
per location $\x$). This specific $B$ image captures peak activity in the
training sequence, and can be efficiently computed as it requires no estimation
of the event PDF; maximum activity is employed as a surrogate for normality.

\begin{figure}[t]
  \centering
\begin{tabular}{cc}
  \footnotesize  Video frame $I_t$ &
  \footnotesize  Motion label field $l_t$\\
  \includegraphics[width=4.0cm]{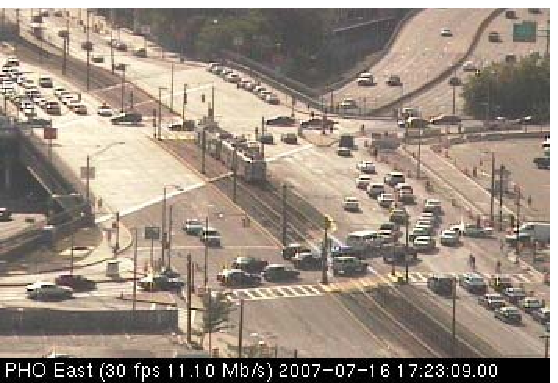}&
  \includegraphics[width=4.0cm]{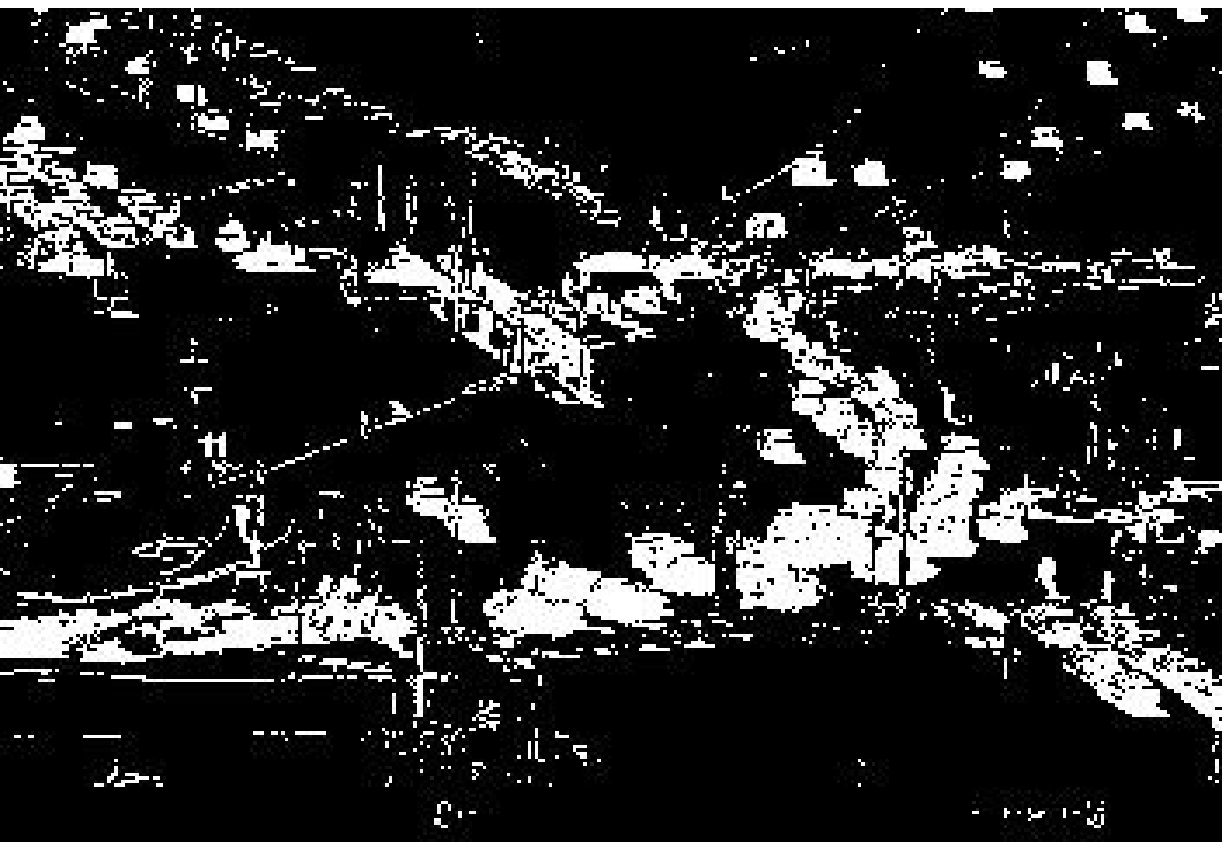}\\

  \footnotesize $B$ image &
  \footnotesize Anomaly map\\
  \includegraphics[width=4.0cm]{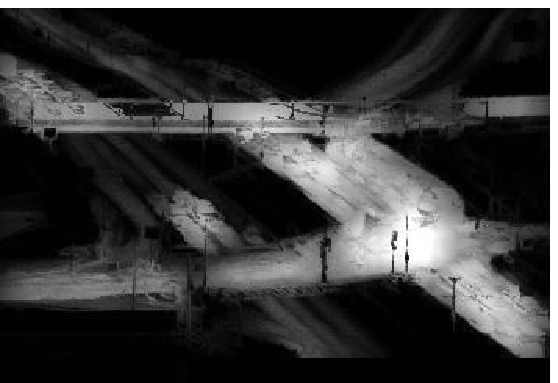}&
  \includegraphics[width=4.0cm]{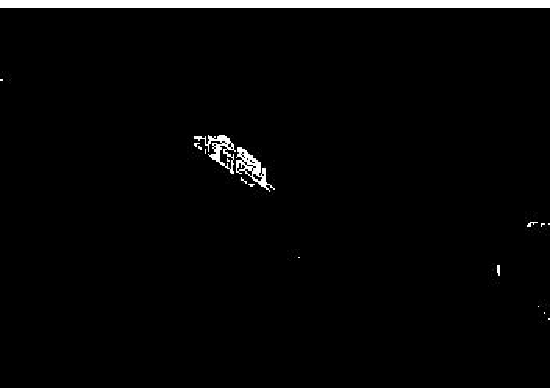}
\end{tabular}
\caption{Behavior subtraction results for the maximum-activity surrogate
  (\ref{eqn:bbehimg1}) on data captured by a stationary, although vibrating,
  camera. This is a highly-cluttered intersection of two streets and interstate
  highway. Although the jitter induces false positives during background
  subtraction ($L_t$), only the tramway is detected by behavior subtraction;
  the rest of the scene is considered normal.}
  \label{fig:Comm}
\vglue -0.4cm
\end{figure}

As shown in Fig.~\ref{fig:Comm}, the $B$ image succinctly synthesizes the
ongoing activity in a training sequence, here a busy urban intersection at peak
hour. It implicitly includes the paths followed by moving objects as well as
the amount of activity registered at every point in the training sequence.

The event model (\ref{eqn:event_prob_suff}) is based on binary random variables
$L$ whose realizations $l$ are computed, for example, using background
subtraction. Since the computed labels $l$ will be necessarily noisy, i.e.,
will include false positives and misses, a positive bias will be introduced
into the event model (even if the noise process is {\it iid}, its mean is
positive since labels $l$ are either $0$ or $1$). The simplest method of noise
suppression is by means of lowpass filtering. Thus, in scenarios with severe
event noise (e.g., unstable camera, unreliable background subtraction) instead
of seeking zero false-alarm rate we opt for event-noise suppression using a
simple averaging filter to compute the background behavior image
\cite{Jodo08icip}:
\begin{eqnarray}\label{eqn:bbehimg2}
  B(\x) = \frac{1}{M} \sum_{t=1}^M e_t(\x).
\end{eqnarray}
%
%
This background behavior image estimates a space-variant bias from the training
data. A non-zero bias can be considered as a temporal stationarity, and
therefore normality, against which observed data can be compared.


%

\subsection{Behavior Subtraction}
\label{ssec:behsub}

Having defined the zero-false-alarm threshold $\tau(0)$ or event-noise bias
{\it via} the background behavior image $B$
(\ref{eqn:bbehimg1}-\ref{eqn:bbehimg2}), we can now apply the event hypothesis
test (\ref{eqn:event_hyptest}) as follows:
\begin{eqnarray*}
  e_t(\x) - B(\x) \decide{abnormal}{normal} \Theta
\end{eqnarray*}
where $\Theta$ is a user-selectable constant allowing for non-zero tolerance
($\Theta=0$ leads to a strict test).
%
%
In analogy to calling $B$ a background behavior image, we call $e_t$ an {\it
  observed behavior image} as it captures events observed in the field of view
of the camera over a window of $w$ video frames. The above test requires the
accumulation of motion labels $l$, object sizes $f$, and state transitions
($\kappa_t$) over $w$ frames. All these quantities can be easily and
efficiently computed.

Clearly, abnormal behavior detection in this case simplifies to the subtraction
of the background behavior image $B$, containing an aggregate of long-term
activity in the training sequence, from the observed behavior image $e_t$,
containing a snapshot of activity just prior to time $t$, and subsequent
thresholding.  This explains the name {\em behavior subtraction} that we gave
to this method.
\begin{figure*}[tb]
  \centering

\begin{tabular}{cccc}
  \footnotesize Video frame $I_t$ &
  \footnotesize Motion label field $l_t$ &
  \footnotesize Object-size descriptor $f_t$ &
  \footnotesize Anomaly map\\
  \includegraphics[width=4.0cm]{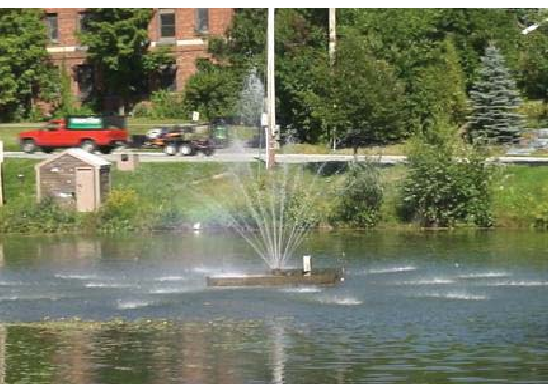}&
  \includegraphics[width=4.0cm]{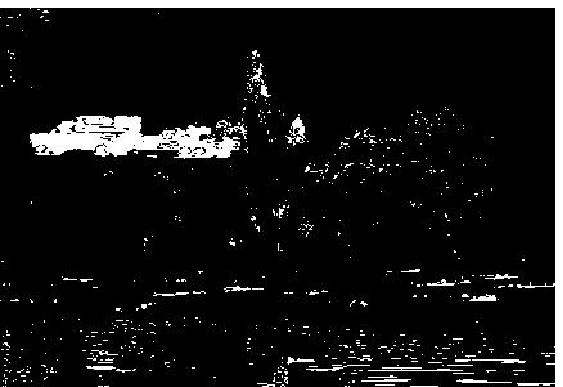}&
  \includegraphics[width=4.0cm]{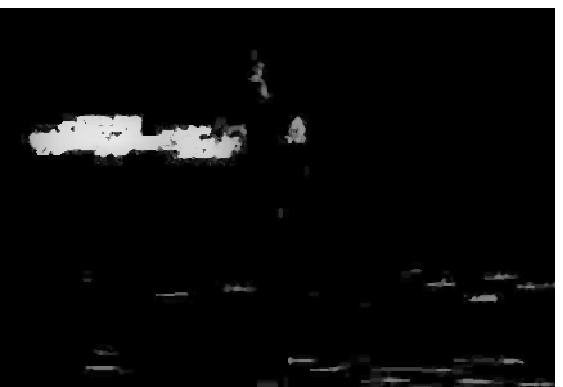}&
  \includegraphics[width=4.0cm]{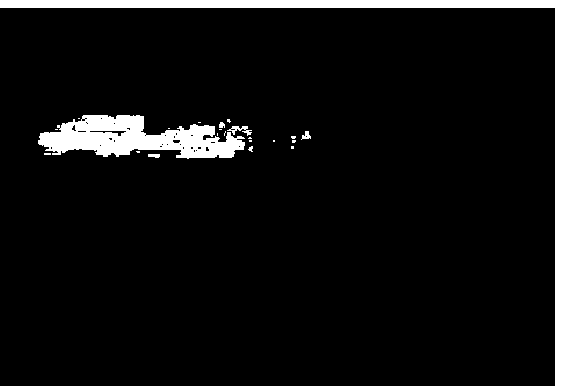}\\

  \includegraphics[width=4.0cm]{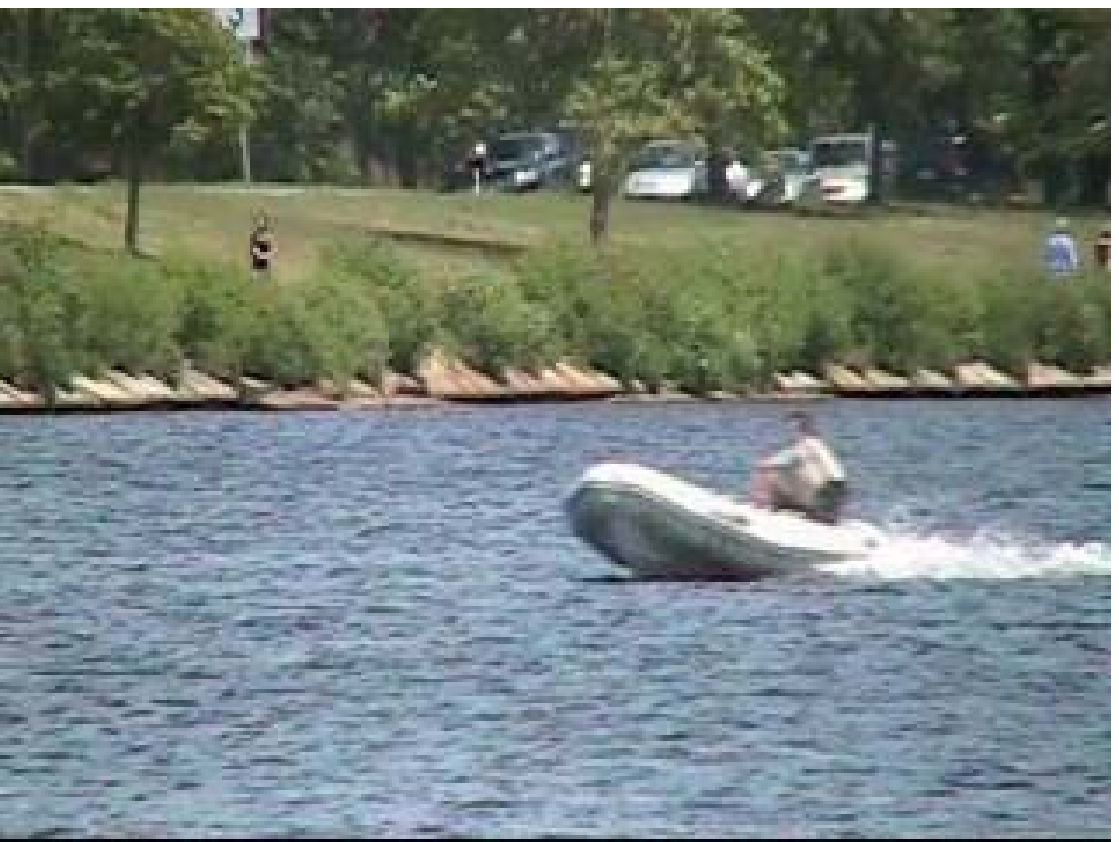}&
  \includegraphics[width=4.0cm]{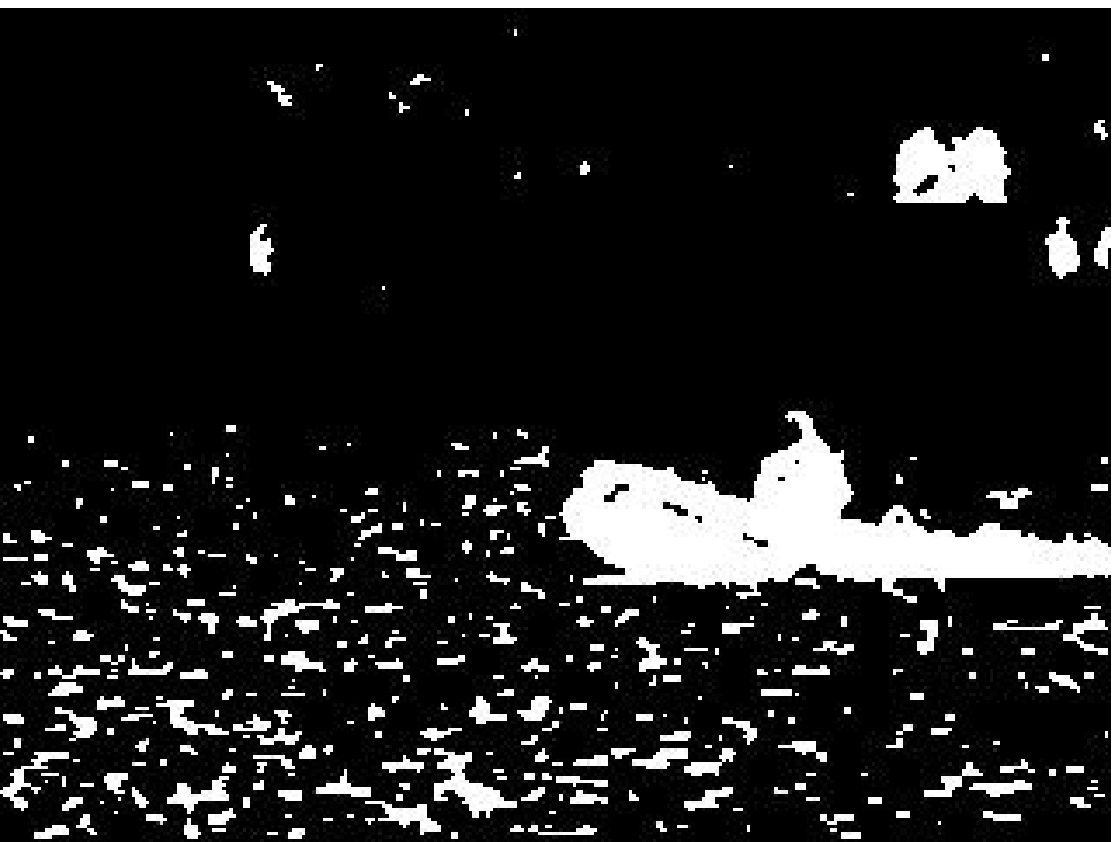}&
  \includegraphics[width=4.0cm]{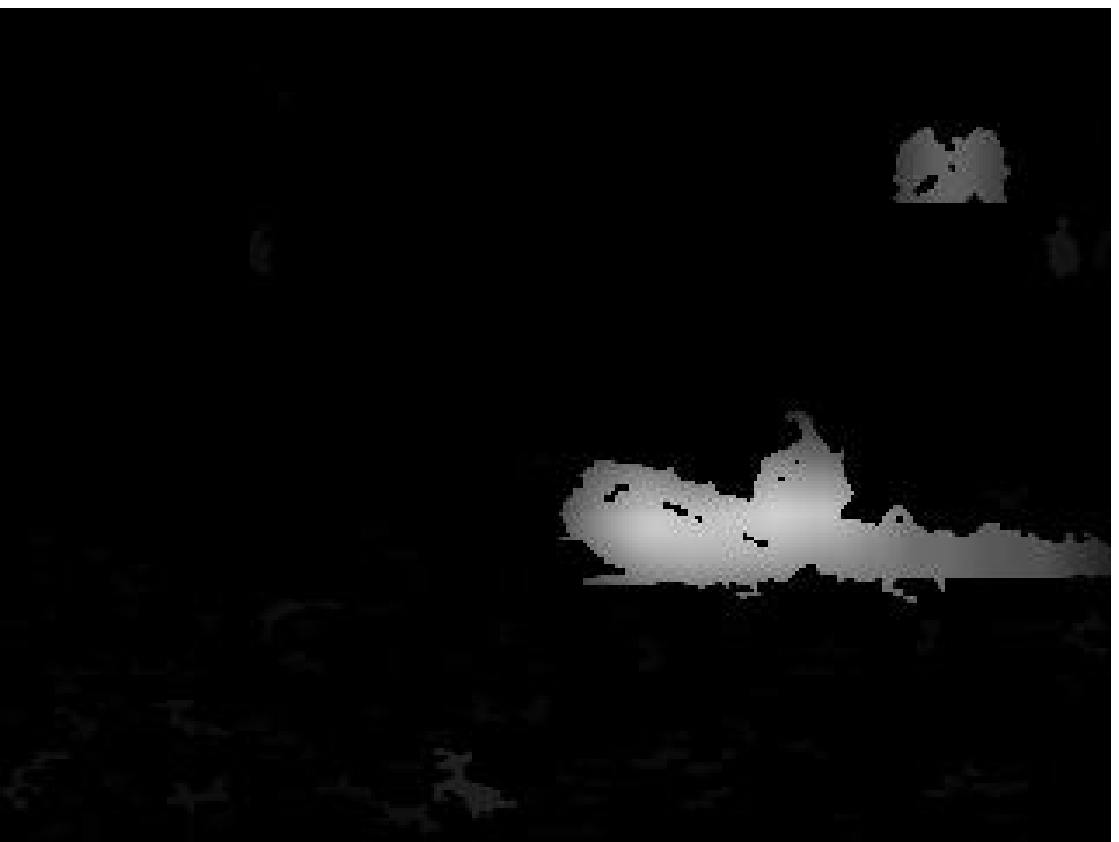}&
  \includegraphics[width=4.0cm]{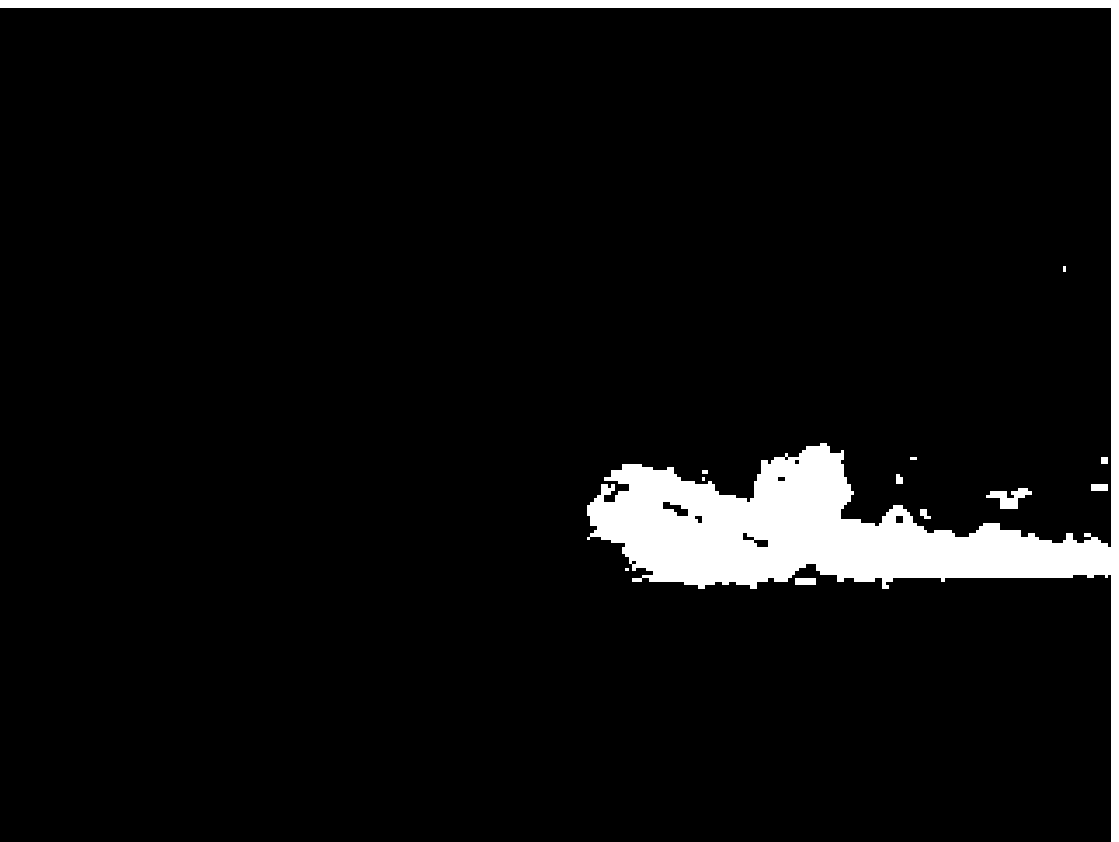}\\

  \includegraphics[width=4.0cm]{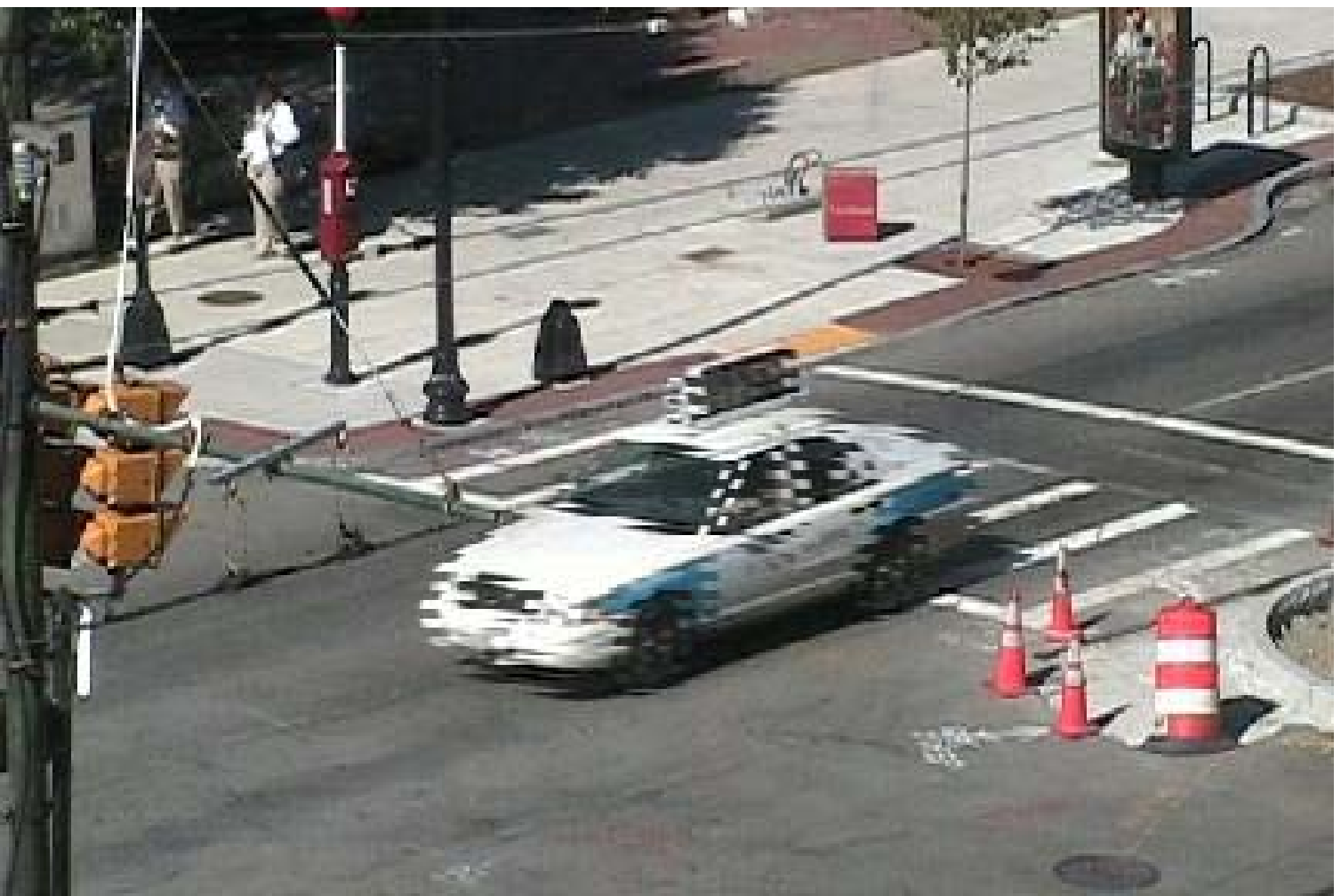}&
  \includegraphics[width=4.0cm]{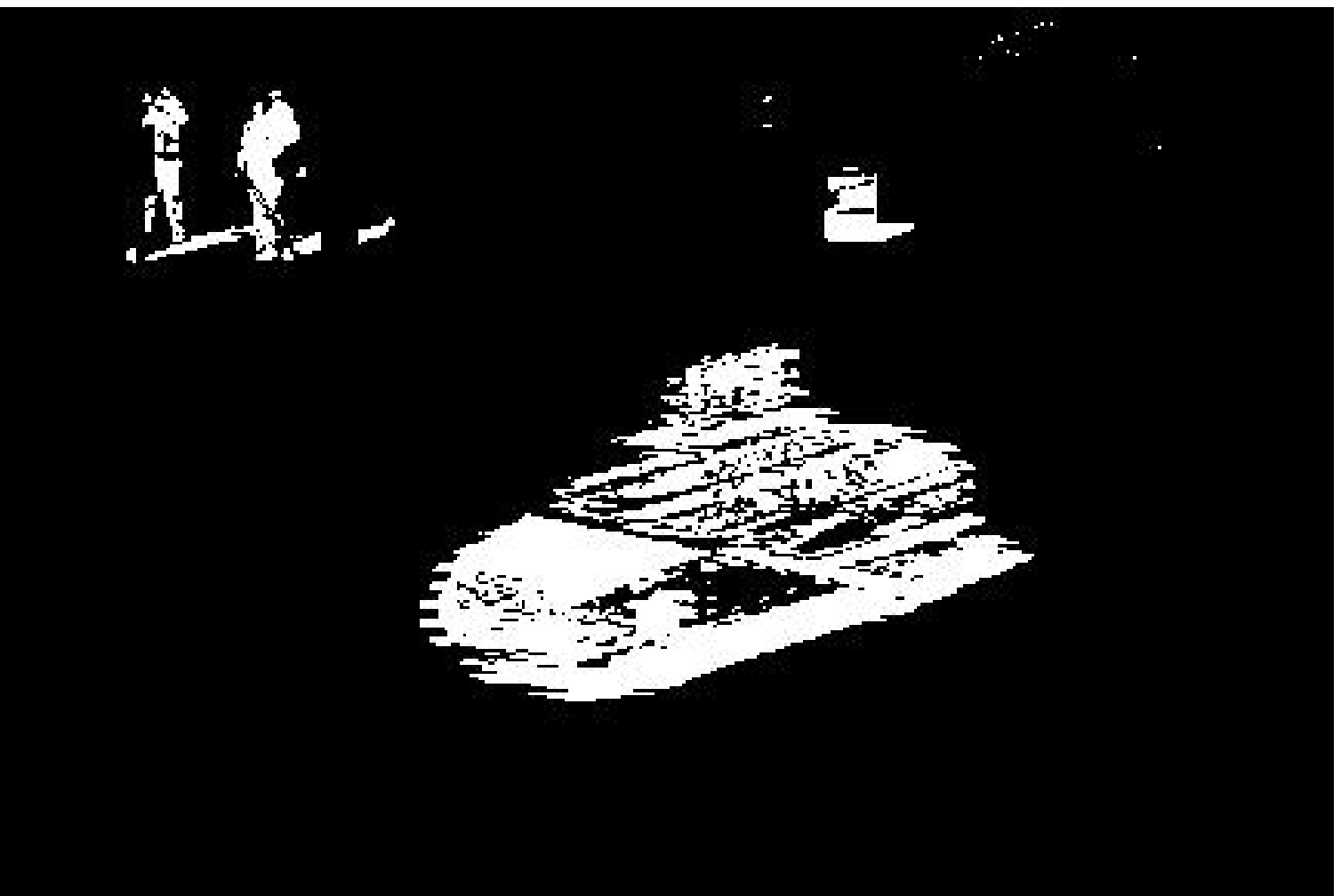}&
  \includegraphics[width=4.0cm]{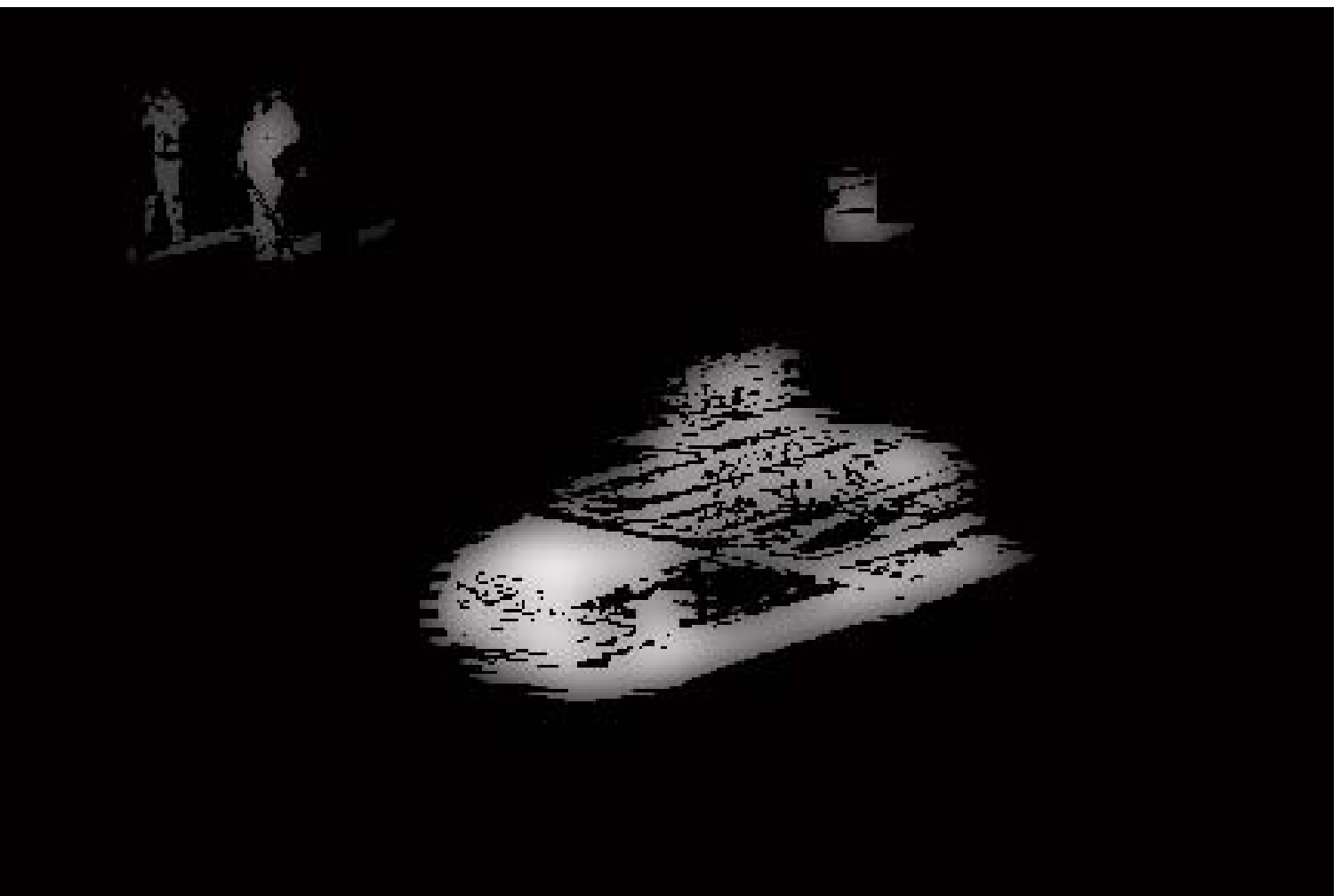}&
  \includegraphics[width=4.0cm]{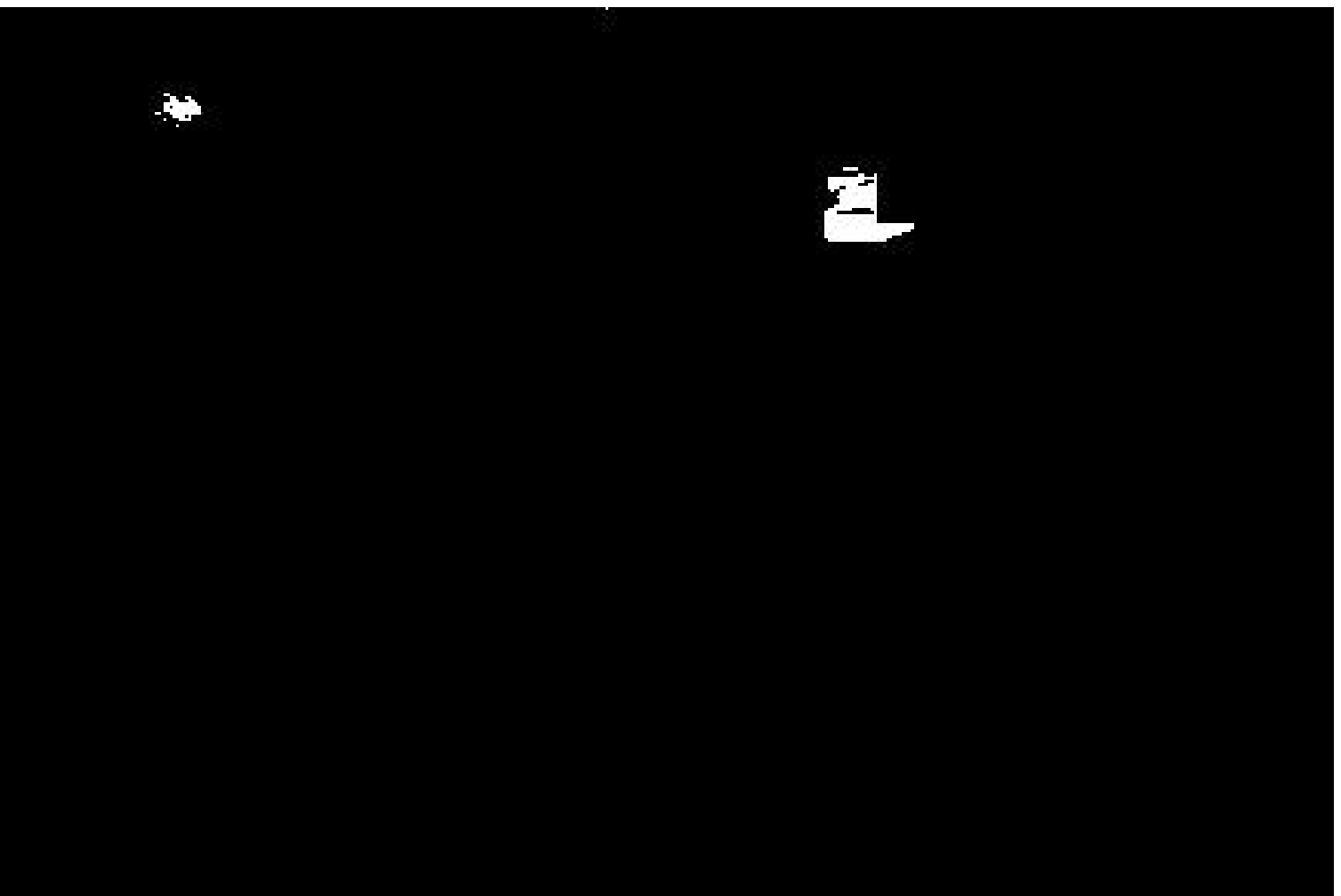}\\

  \includegraphics[width=4.0cm]{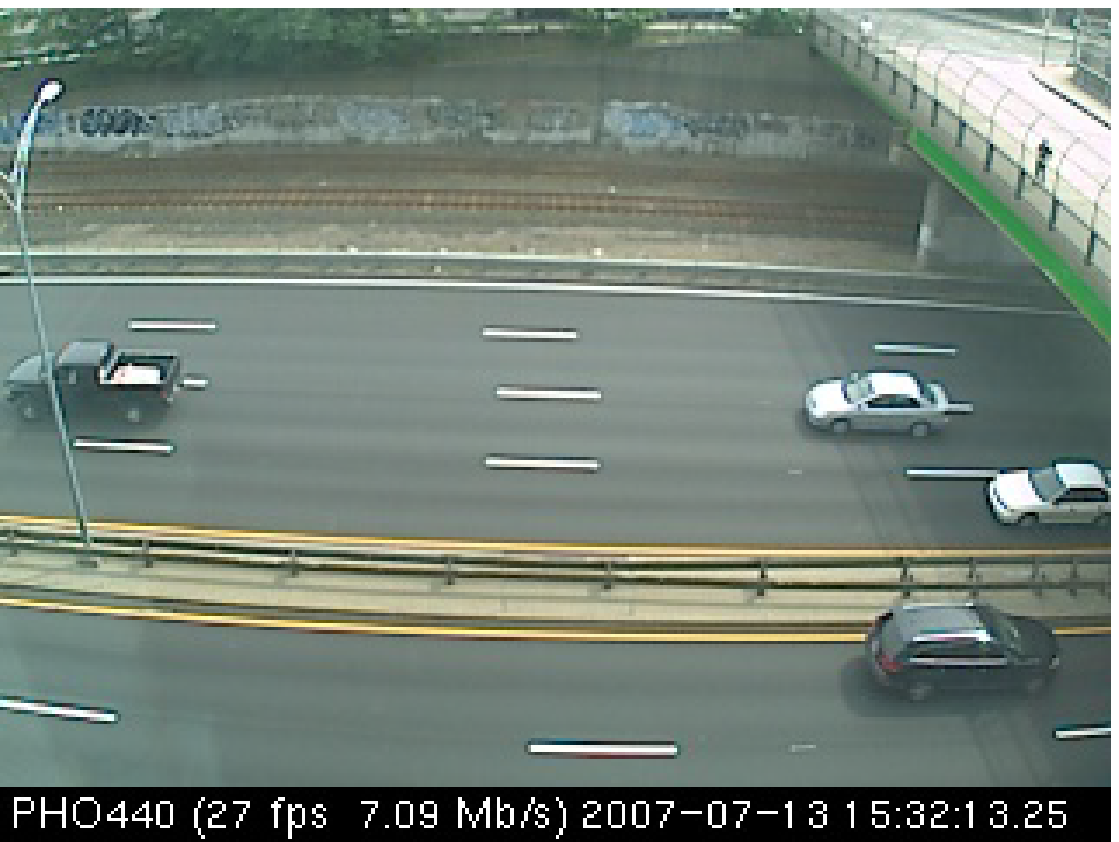}&
  \includegraphics[width=4.0cm]{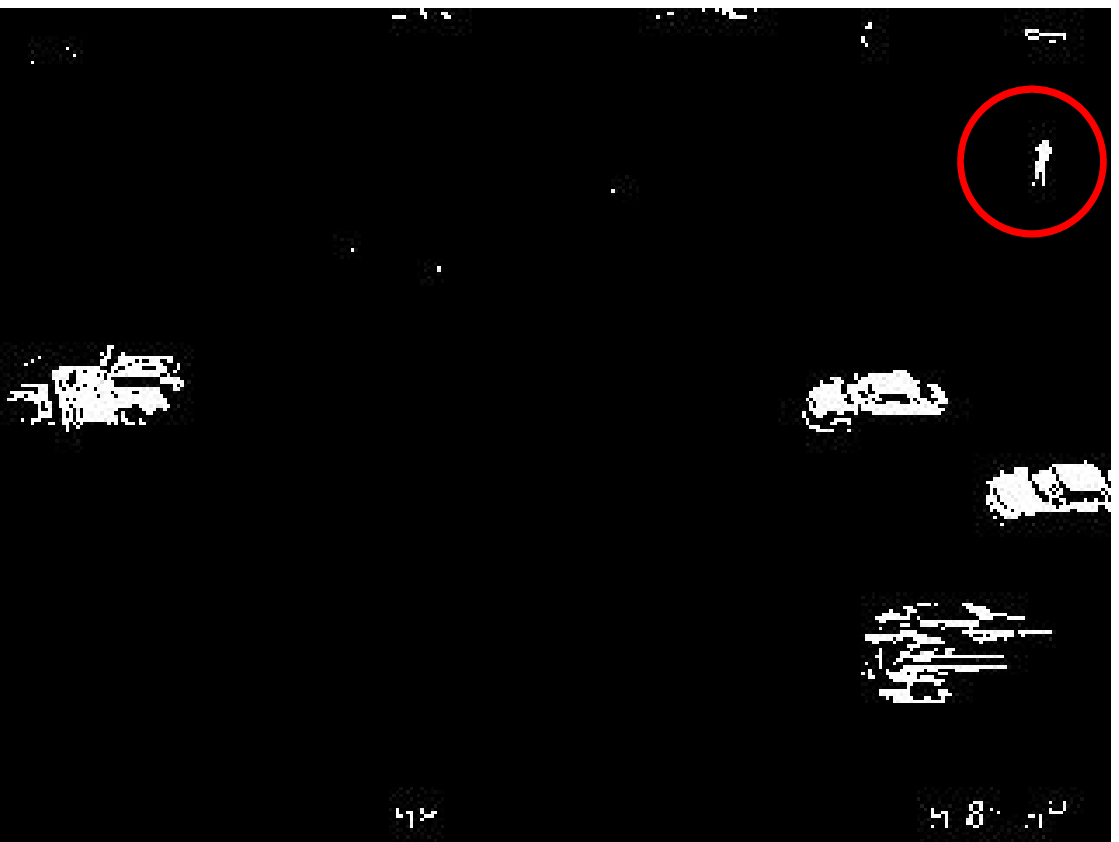}&
  \includegraphics[width=4.0cm]{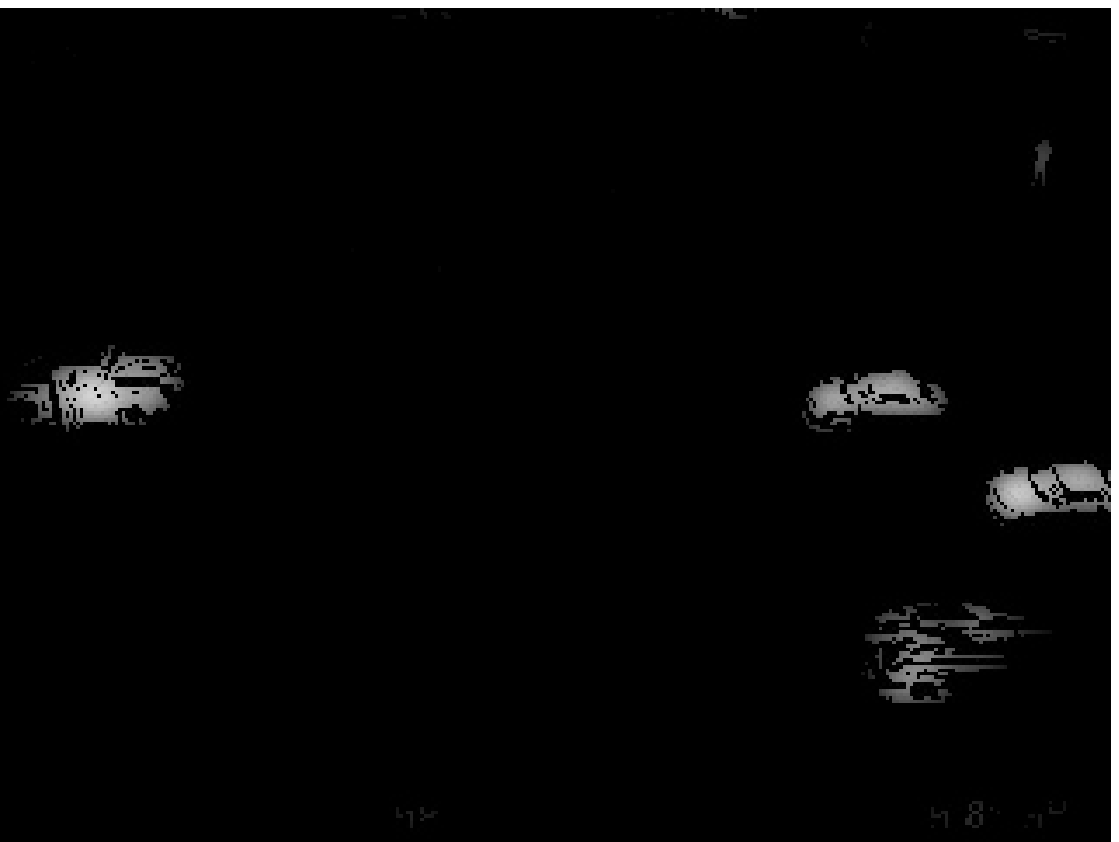}&
  \includegraphics[width=4.0cm]{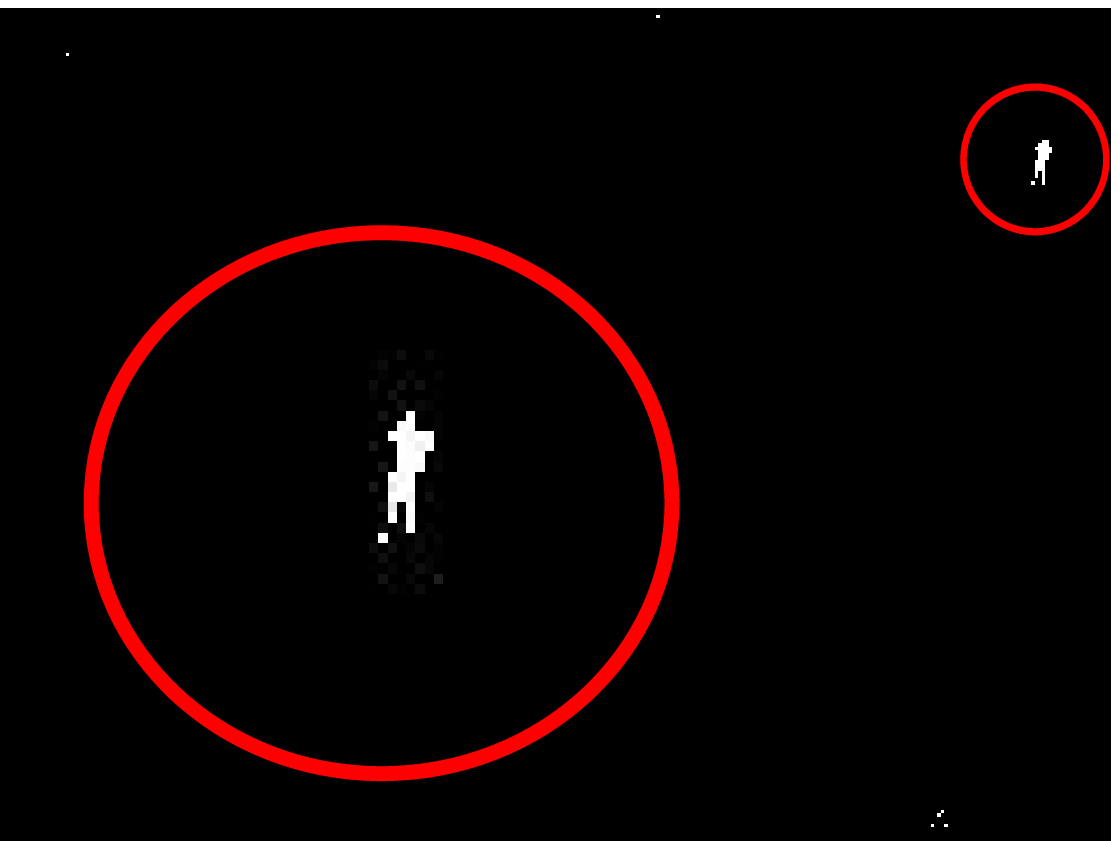}
\end{tabular}
\caption{Behavior subtraction results for maximum-activity surrogate
  (\ref{eqn:bbehimg1}) on video sequences containing shimmering water surface
  (two top rows), strong shadows (third row) and very small abnormally-behaving
  object (bottom row).}
  \label{fig:robustness}
\vglue -0.6cm
\end{figure*}

\section{Experimental Results}
\label{sec:expres}

We tested our behavior subtraction algorithm for both the maximum- and
average-activity surrogates on black-and-white and color, indoor and outdoor,
urban and natural-environment video sequences. In all cases, we computed the
label fields $l_t$ using simple background subtraction (\ref{eq:md}) with $\tau
= 40$ and background $b$ updated with $\alpha$ between $10^{-3}$ and $10^{-2}$,
depending on the sequence. Although we have performed experiments on a wide
range of model parameters, we are presenting here the results for event model
based on size descriptor (\ref{eqn:motobjdes}) ($A_1=A_2=0$).

\begin{figure}[t]
  \centering
\begin{tabular}{cc}
  \footnotesize Video frame $I_{t=0}$ &
  \footnotesize Video frame $I_{t=2240}$ \\
  \includegraphics[width=4.0cm]{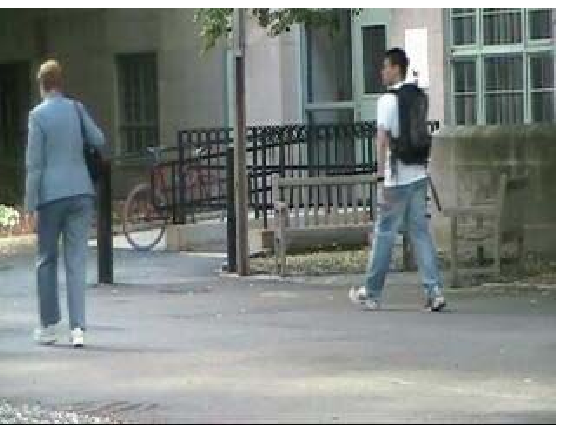}&
  \includegraphics[width=4.0cm]{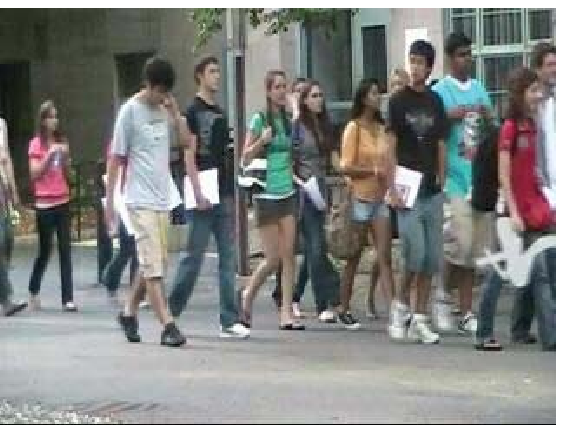}\\

  \footnotesize Object-size descriptor $f_{t=0}$ &
  \footnotesize Object-size descriptor $f_{t=2240}$ \\
  \includegraphics[width=4.0cm]{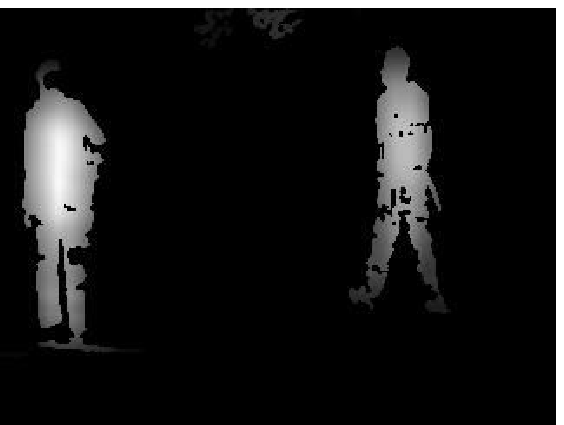}&
  \includegraphics[width=4.0cm]{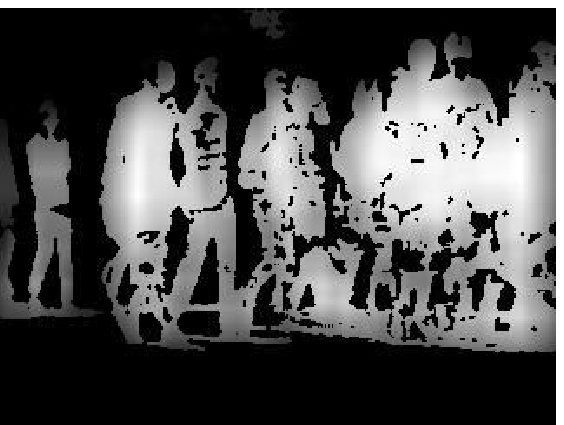}\\

  \footnotesize Anomaly map at $t=0$ &
  \footnotesize Anomaly map at $t=2240$ \\
  \includegraphics[width=4.0cm]{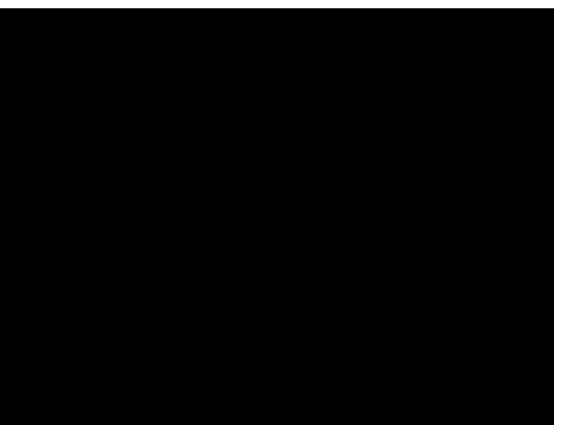}&
  \includegraphics[width=4.0cm]{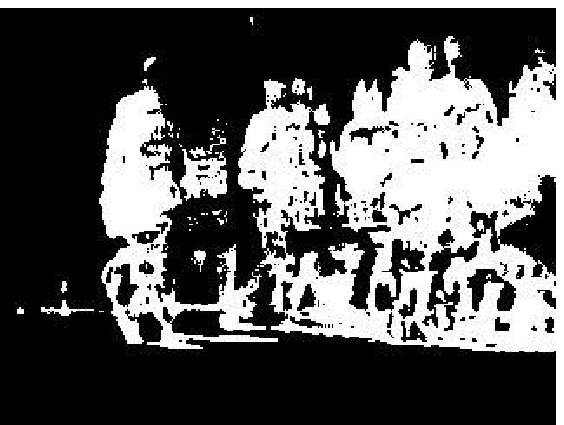}
\end{tabular}
\caption{Results of behavior subtraction for the maximum-activity surrogate
  (\ref{eqn:bbehimg1}) with training performed on a video containing single
  pedestrian and fluttering leaves (top of the frame).  The group of
  pedestrians is associated with a large amount of activity and thus detected
  by our method as anomaly.}
\label{fig:groupCampus}
\vglue -0.6cm
\end{figure}

The results of behavior subtraction using the maximum-activity surrogate
(\ref{eqn:bbehimg1}) are shown in Figs.~\ref{fig:Comm}-\ref{fig:groupCampus}.
Each result was obtained using a training sequence of length $M$=1000-5000
frames, $w=100$, and $\Theta \in [0.5, 0.7]$. As is clear from the figures, the
proposed method is robust to inaccuracies in motion labels $l_t$.  Even if
moving objects are not precisely detected, the resulting anomaly map is
surprisingly precise. This is especially striking in Fig.~\ref{fig:Comm} where
a highly-cluttered environment results in high density of motion labels while
camera jitter corrupts many of those labels.

Behavior subtraction is also effective in removal of unstructured, parasitic
motion such as due to water activity (fountain, rain, shimmering surface), as
illustrated in Fig.~\ref{fig:robustness}. Note that although motion label
fields $l_t$ include unstructured detections due to water droplets, only the
excessive motion is captured by the anomaly maps (passenger car and truck with
trailer). Similarly, the shimmering water surface is removed by behavior
subtraction producing a fairly clean boat outline in this difficult scenario.
Our method also manages to detect abandoned objects and people lingering, as
seen in the two bottom rows of Fig.~\ref{fig:robustness}.

Fig.~\ref{fig:groupCampus} shows yet another interesting outcome of behavior
subtraction. In this case the background behavior image was trained on a video
with single pedestrian and fluttering leaves. While the object-size descriptor
captures both individual pedestrians and groups thereof, anomalies are detected
only when a large group of pedestrians passes in front of the camera.

\begin{figure}[t]
  \centering
\begin{tabular}{cc}
  \footnotesize Video frame $I_t$ &
  \footnotesize Single-Gaussian method \\
  \includegraphics[width=4.0cm]{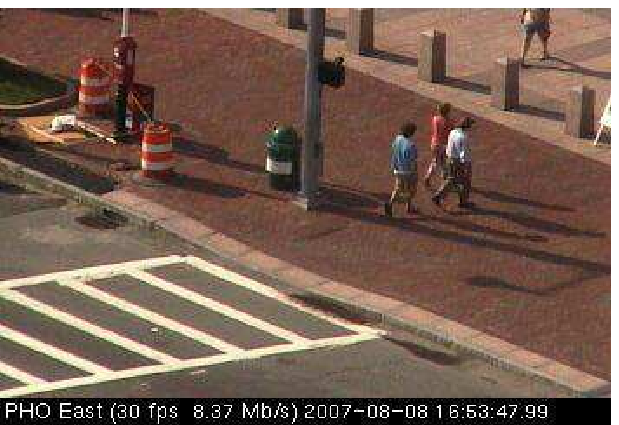} &
  \includegraphics[width=4.0cm]{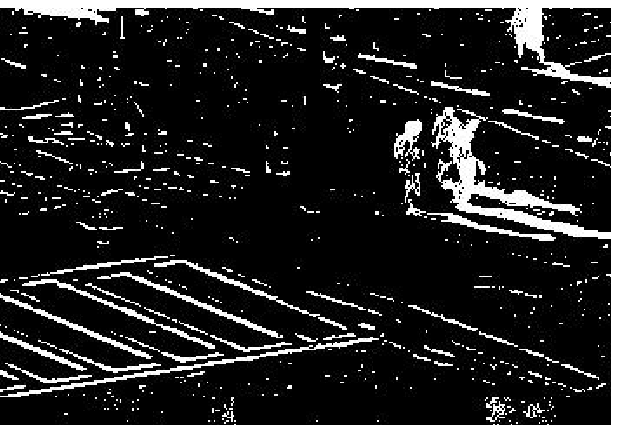} \\

  \footnotesize Parzen-window method &
  \footnotesize Behavior subtraction \\
  \includegraphics[width=4.0cm]{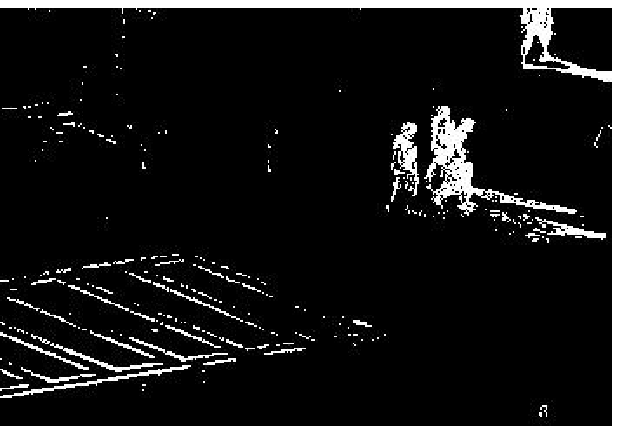} &
  \includegraphics[width=4.0cm]{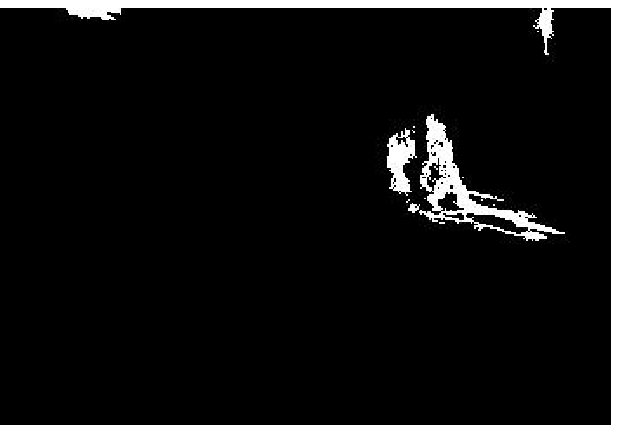}
\end{tabular}
\caption{Results for background subtraction based on single-Gaussian
  \cite{Wren97} and non-parametric-kernel \cite{Elga02} hypothesis tests,
  as well as for behavior subtraction, on data captured by severely vibrating
  camera. Camera jitter introduces excessive false positives in both background
  subtraction methods while behavior subtraction is relatively immune to
  jitter.}
\label{fig:motiondetection}
\vglue -0.6cm
\end{figure}

The results of behavior subtraction using the average-activity surrogate are
shown in Fig.~\ref{fig:motiondetection}.  The video sequence has been captured
by a vibrating camera (structural vibrations of camera mount).  It is clear
that behavior subtraction with average-activity surrogate outperforms
background subtraction based on single-Gaussian model \cite{Wren97} and
non-parametric-kernel model \cite{Elga02}. As can be seen, behavior subtraction
effectively eliminates false positives without significantly increasing misses.

As already mentioned, the proposed method is efficient in terms of processing
power and memory use, and thus can be implemented on modest-power processors
(e.g., embedded architectures). For each pixel, it requires one floating-point
number for each pixel of $B$ and $e$, and $w/8$ bytes for $l$.  This
corresponds to a total of 11 bytes per pixel for $w=24$.  This is significantly
less than 12 floating-point numbers per pixel needed by a tri-variate Gaussian
for color video data (3 floating-point numbers for $R,G,B$ means and 9 numbers
for covariance matrix).
%
%
%
%
Our method currently runs in {\em Matlab} at 20 fps on $352\times 240$-pixel
video using a 2.1 GHz dual-core Intel processor. More experimental results can
be found in our preliminary work \cite{Jodo08vcip,Jodo08icip}, while complete
video sequences can be downloaded from {\small\tt
  www.dmi.usherb.ca/$\sim$jodoin/projects/PAMI\_2009}.

\section{Conclusions}
\label{sec:concl}

In this paper, we proposed a framework for the characterization of dynamic
events and, more generally, behavior. We defined events as spatio-temporal
signatures composed of various moving-object features, and modeled them using
stationary random processes. We also proposed a computationally-efficient
implementation of the proposed models, called behavior subtraction. In fact,
due to simple surrogates of activity/behavior statistics used, behavior
subtraction is very easy to implement, uses little memory and can run on an
embedded architecture. Furthermore, the proposed framework is content-blind,
i.e., equally applicable to pedestrians, motor vehicles or animals. Among
applications that can benefit from the proposed framework are suspicious
behavior detection and motion detection in presence of strong parasitic
background motion. Yet, challenges remain. One challenge is to extend the
proposed concepts to multiple cameras so that a mutual reinforcement of
decisions takes place; some of our preliminary work can be found in
\cite{Ermi08icdsc}. Another challenge is to detect anomalies at object level
while using only pixel-level decisions proposed here.


\end{document}